\documentclass{article}
\usepackage{proceed2e}
\usepackage{times}
\usepackage{chicagor}
\usepackage{graphicx} %
\usepackage{times}
\usepackage{color}
\usepackage{subfigure}

\belowdisplayskip \abovedisplayskip

\newlength{\sectionReduceTop}
\newlength{\sectionReduceBot}
\newlength{\subsectionReduceTop}
\newlength{\subsectionReduceBot}
\newlength{\abstractReduceTop}
\newlength{\abstractReduceBot}
\newlength{\captionReduceTop}
\newlength{\captionReduceBot}
\newlength{\subsubsectionReduceTop}
\newlength{\subsubsectionReduceBot}

\newlength{\horSkip}
\newlength{\verSkip}

\newlength{\figureHeight}
\newtheorem{THEOREM}{Theorem}[section]
\newenvironment{theorem}{\begin{THEOREM} \hspace{-.85em} {\bf :} }%
                        {\end{THEOREM}}
\newtheorem{LEMMA}[THEOREM]{Lemma}
\newenvironment{lemma}{\begin{LEMMA} \hspace{-.85em} {\bf :} }%
                      {\end{LEMMA}}
\newtheorem{COROLLARY}[THEOREM]{Corollary}
\newenvironment{corollary}{\begin{COROLLARY} \hspace{-.85em} {\bf :} }%
                          {\end{COROLLARY}}
\newtheorem{PROPOSITION}[THEOREM]{Proposition}
\newenvironment{proposition}{\begin{PROPOSITION} \hspace{-.85em} {\bf :} }%
                            {\end{PROPOSITION}}
\newtheorem{DEFINITION}[THEOREM]{Definition}
\newenvironment{definition}{\begin{DEFINITION} \hspace{-.85em} {\bf :} \rm}%
                            {\end{DEFINITION}}
\newtheorem{CLAIM}[THEOREM]{Claim}
\newenvironment{claim}{\begin{CLAIM} \hspace{-.85em} {\bf :} \rm}%
                            {\end{CLAIM}}
\newtheorem{EXAMPLE}[THEOREM]{Example}
\newenvironment{example}{\begin{EXAMPLE} \hspace{-.85em} {\bf :} \rm}%
                            {\end{EXAMPLE}}
\newtheorem{REMARK}[THEOREM]{Remark}
\newenvironment{remark}{\begin{REMARK} \hspace{-.85em} {\bf :} \rm}%
                            {\end{REMARK}}

\newcommand{\thm}{\begin{theorem}}
\newcommand{\lem}{\begin{lemma}}
\newcommand{\pro}{\begin{proposition}}
\newcommand{\dfn}{\begin{definition}}
\newcommand{\rem}{\begin{remark}}
\newcommand{\xam}{\begin{example}}
\newcommand{\cor}{\begin{corollary}}
\newcommand{\prf}{\noindent{\bf Proof:} }
\newcommand{\ethm}{\end{theorem}}
\newcommand{\elem}{\end{lemma}}
\newcommand{\epro}{\end{proposition}}
\newcommand{\edfn}{\bbox\end{definition}}
\newcommand{\erem}{\bbox\end{remark}}
\newcommand{\exam}{\bbox\end{example}}
\newcommand{\ecor}{\end{corollary}}
\newcommand{\eprf}{\bbox\vspace{0.1in}}
\newcommand{\beqn}{\begin{equation}}
\newcommand{\eeqn}{\end{equation}}

\newcommand{\bbox}{\vrule height7pt width4pt depth1pt}

\newcommand{\clm}{\begin{claim}}
\newcommand{\eclm}{\end{claim}}

\newcommand{\inter}{\cap}

\newcommand{\IR}{\mbox{$I\!\!R$}}

\renewcommand{\phi}{\varphi}

\newcommand{\ol}{\setlength{\itemsep}{0pt}\begin{enumerate}}
\newcommand{\eol}{\end{enumerate}\setlength{\itemsep}{-\parsep}}
\newcommand{\ul}{\setlength{\itemsep}{0pt}\begin{itemize}}
\newcommand{\dl}{\setlength{\itemsep}{0pt}\begin{description}}
\newcommand{\edl}{\end{description}\setlength{\itemsep}{-\parsep}}
\newcommand{\eul}{\end{itemize}\setlength{\itemsep}{-\parsep}}

\newcommand{\commentout}[1]{}

\newcommand{\bi}{\begin{itemize}}
\newcommand{\ei}{\end{itemize}}
\newcommand{\be}{\begin{enumerate}}
\newcommand{\ee}{\end{enumerate}}

\renewcommand{\dfn}{\begin{definition}}
\renewcommand{\edfn}{\end{definition}}
\renewcommand{\xam}{\begin{example}}
\renewcommand{\exam}{\end{example}}
\renewcommand{\thm}{\begin{theorem}}
\renewcommand{\ethm}{\end{theorem}}
\renewcommand{\cor}{\begin{corollary}}
\renewcommand{\ecor}{\end{corollary}}
\renewcommand{\lem}{\begin{lemma}}
\renewcommand{\elem}{\end{lemma}}
\renewcommand{\pro}{\begin{proposition}}
\renewcommand{\epro}{\end{proposition}}

\newcommand{\leaveout}[1]{}
\newcommand{\fullv}[1]{#1}
\newcommand{\shortv}{\commentout}
\newcommand{\othm}[1]{\begin{oldthm}{\ref{#1}}}
\newcommand{\eothm}{\end{oldthm} \medskip}
\newenvironment{oldthm}[1]{\par\noindent{\bf Theorem #1:} \em \noindent}{\par}
\renewcommand{\cite}{\shortcite}

\begin{document}
\title{MDPs with Unawareness in Robotics}
\author{ Nan Rong \ \ \ \ Joseph Y. Halpern \ \ \ \ Ashutosh Saxena\\
Computer Science Department, Cornellx University\\
Ithaca, NY 14853\\
$\{$rongnan $|$ halpern $|$ asaxena$\}$@cs.cornell.edu
}
\maketitle

\begin{abstract}
We formalize decision-making problems in robotics and automated control
using continuous MDPs and actions that take place over continuous
time intervals.  We then approximate the continuous MDP using finer and
finer discretizations.  Doing this results in
a family of systems, each of which has an extremely large action
space, although only a few  actions are ``interesting''.
We can view the decision maker as being unaware of which actions
are  ``interesting''.    
We can model this using 
\emph{MDPUs}, MDPs with unawareness, 
where the action space is much smaller.
As we show, MDPUs can be used as a general framework for  learning
tasks in robotic problems. We prove results on  
the difficulty 
of learning a near-optimal policy in an an MDPU for a continuous task.
We apply these ideas to the problem of having a humanoid robot learn
on its own how to walk.
\end{abstract}

\section{INTRODUCTION}\label{sec:intro}
Markov decision processes (MDPs) are widely used for modeling decision
making problems in robotics and automated control.
Traditional MDPs assume that the decision maker (DM) knows all states and 
actions. 
However, in many robotics applications, %
the space of states and actions is continuous.
To find appropriate policies, we typically discretize both states and
actions.  However, we do not know in advance what level of
discretization is good enough for getting a good policy.  Moreover, in
the discretized space, the set of actions is huge.  However, relatively
few of the actions are ``interesting''.
For example, when flying a robotic helicopter, only a small set of
actions lead to useful flying techniques; an autonomous helicopter must
learn these techniques.  
Similarly, a humanoid robot needs to learn various  maneuvers (e.g., walking or running) that enable it to move around, but the space of potential actions that it must search to find a successful gait is huge, while most actions result in the robot losing control and falling down.

Halpern, Rong, and Saxena \citeyear{HRS10} (HRS from now on)
defined \emph{MDPs with
unawareness 
(MDPUs)}, where a decision-maker (DM) can be unaware of the actions in an MDP.
In the robotics applications in which we are interested, we
can think of the DM
(e.g., a humanoid robot)
as being unaware of which actions are the useful
actions, and thus can model what is going on using an MDPU. 

In this paper, we apply MDPUs to continuous problems. 
We model  such problems using \emph{continuous MDPs}, 
where actions are performed over a continuous duration of time.  
Although many problems fit naturally in our continuous MDP framework,
and there has been a great deal of work on continuous-time MDPs, our
approach seems new, and of independent interest.  (See the discussion in
Section~\ref{sec:related}.)  
It is hard to find near-optimal policies in continuous MDPs.
A standard approach is
to use discretization.
We use discretization as well, but our discrete models are MDPUs, rather
than MDPs, which allows us both to use relatively few actions (the
``interesting actions''), while taking into account the possibility of
there being interesting actions that the DM has not yet discovered.
We would like to find a discretization 
level for which the optimal policy in the MDP underlying the
approximating MDPU provides a good approximation to the
optimal policy in the continuous MDP that accurately describes the
problem, and then find a near-optimal policy in that discretized MDPU.

HRS gave a complete
characterization of when it is possible
to learn to play near-optimally in an MDPU, extending earlier work
\cite{BT02,KS02} showing that it is always possible to learn to play
near-optimally in an MDP.  
We extend and generalize these results so as to apply them to the
continuous problems of interest to us.  
We characterize when brute-force exploration can be used to find a near-optimal policy in our setting,
and show that a variant of the URMAX
algorithm presented by HRS can find a near-optimal policy. 
We also characterize the complexity of learning to play near-optimally
in continuous problems, when more ``guided'' exploration is used.
Finally, we discuss how MDPUs can be used to solve a real
robotic problem: to enable a humanoid robot to learn walking on its
own. In our experiment, the robot learned various gaits at multiple
discretization levels, including both forward and backward gaits; both
efficient and inefficient gaits; and both gaits that resemble human
walking, and those that do not. 
\commentout{
However, brute-force is clearly quite inefficient.  
Our results provide some insight into when 
human assistance is needed to learn a near-optimal policy,
as was the case for Abbeel and Ng \citeyear {AN05ICML} and Kober and
Peters \citeyear{KP08}. 
Finally, we  discuss how these results were used to guide the development
of a robotic remote control (RC) car that can perform drifting actions.
\commentout{
We show here that these results can help us understand when brute-force
search can be used to learn near-optimal policies in robotics applications.
These results help explain why, in papers such as \cite{AN05ICML,KP08},
human assistance was needed to help the robot to learn a good policy;

we also used them to guide the development of a near-optimal policy to
learn drifting in a robotic car.
}
}%

\section{MDPU: A REVIEW}\label{sec:mdpu}
\fullv{In this section, we review the definition of MDPU and the results of
HRS show that
is possible for a DM to learn to play near-optimally.}
\fullv{In the standard MDP model, we have a set of actions, and states
representing the complete set of actions and states that are available
to the Decision Maker (DM).} 
Describing a situation by a standard MDP misses out on some
important features.
In general, an agent may not be aware of all the actions that can be
performed.  For example, an agent playing a video game may not be aware
of all actions that can be performed in a given state.  
Our model is compatible with a number of interpretations of unawareness.
In the robotics setting, we take a particular concrete interpretation.  
Here, the
number of actions is typically extremely large,
but only a few of these actions are actually useful. For example,
although an autonomous 
helicopter can have a huge number of actions, only a few are useful
for flying the helicopter, whereas the rest simply result in a crash.
We can abstract what is going on by defining a set of
``useful actions''.  The DM may initial be unaware of many or even most
of the useful actions.  A DM may become aware of a useful action (and
thus, can safely perform it) by performing a special action called
the  \emph{explore} action, denoted $a_0$.
Playing the explore action results in the DM learning about new actions
with some probability.

We thus take an MDPU to be a tuple
$M=(S,A,A_0,g,a_0,g_0,P,D,R)$,\footnote{
The MDPU model in HRS also includes $R_i^+$ and $R_i^-$,
which are the reward (resp., penalty) functions for playing
$a_0$ and  discovering (resp., not discovering) a useful 
action. We omit them here; they play no role in 
the theorems that we are citing, and would only clutter our later
presentation.  
} where the tuple
$(S,A,g,P,R)$ is a standard MDP---that is, $S$ is a set of states, $A$ is
a set of actions, $g(s)$ is the set of actions available at state $s$,
for each tuple $(s,s',a) \in S \times S
\times A$, $P(s,s',a)$ gives the probability of making a transition from
$s$ to $s'$ if action $a$ is performed, and $R(s,s',a)$ gives the reward
earned by the DM if this transition is taken;
$A_0\subseteq A$ is the set of actions that the DM initially knows
to be useful; 
$a_0$ is the special 
\textit{explore} action; 
$g(s) \subseteq A$ is the set of actions that can be performed at state $s$;
$g_0(s) \subseteq A_0 \inter g(s)$ is the set of actions that the DM is
aware of at state $s$; 
finally, $D$ is the \textit{discovery probability
function}. $D(j,t,s)$ is the probability of discovering a useful action
given that there are $j$ useful actions to be discovered at state $s$,
and $a_0$ has 
already been played $t-1$ times without discovering a useful
action. Intuitively, 
$D$ describes how quickly the DM can discover a useful action. We assume
that $D(j,t,s)$ is non-decreasing as a function of $j$: the more useful
actions there are to be found, the easier we can find one. How
$D(j,t,s)$ varies with $t$ depends on 
the problem. In the sequel, we assume for ease of exposition that $D(j,t,s)$ is
independent of $s$, so we write $D(j,t)$ rather than $D(j,t,s)$.
$M' = (S,A,P,R)$ is called the \emph{MDP underlying $M$}.

Kearns and Singh \citeyear{KS02} and Brafman and Tennenholtz
\citeyear{BT02} have studied the problem of learning how to play 
near-optimally in an MDP.
Roughly speaking, to play near-optimally means that, for all $\epsilon>0$ and
$\delta>0$, we can find a policy that obtains
 expected reward $\epsilon$-close to that of the optimal policy with
probability at least $1-\delta$.
HRS completely characterize the
difficulty of learning to play near-optimally in an MDPU.
We briefly review the relevant results here.  
\commentout{
There are three
theorems, all of which discuss the difficulty of learning to
play near-optimally in an MDPU. %
\begin{enumerate}
\item \textbf{Impossibility Theorem}: This theorem discusses when it is impossible 
for the DM to use computational power to learn a near-optimal policy.
\item \textbf{General Possibility Theorem}: This theorem discusses when it is 
possible for the DM to use computational power to learn 
a near-optimal policy.
\item \textbf{Polynomial and Exponential Time Cases}: This corollary discusses 
the specific case, where the computational time required for learning
an optimal policy is polynomial and exponential respectively.
\end{enumerate}
}
Despite initially being unaware of some
useful actions, we want the DM to learn a policy that is
near-optimal in the underlying MDP.
HRS showed that whether a DM can learn to play optimally and how long it takes 
depend on the value of $D(1,t)$---the probability of discovering a new
action given that there is a new action to discover and the DM has tried
$t-1$ times in the past to discover a new action.
The first result characterizes when it is impossible to
learn to play near-optimally.  
\fullv{
It turns out that the result holds even
if the DM has quite a bit of information, as made precise by the
following definition.

\dfn\label{d0}
Define a DM to be \textit{quite knowledgeable} if (in addition to $S$, $D$)
she knows $|A|$, the transition
function $P_0$, the reward function $R_0$ for states in $S$ and
actions in $A_0$, and $R_{\max}$.
\edfn

Another relevant concept that we need to define is the mixing time. 
}%
\shortv{  
To make it precise, we recall the notion of mixing time.
}%
\fullv{
A policy $\pi$ may take a long time to reach its expected 
payoff. For example, if getting a high reward involves reaching a particular state
$s^*$, and the probability of reaching $s^*$ from some state $s$ is low,
then the time to get the high reward will be high.  To deal with this,
Kearns and Singh \citeyear{KS02} argue that the running time of a
learning algorithm should be compared to the time that an algorithm 
with full information takes to get a comparable reward.}
\fullv{
\dfn\label{d1}
Define the \emph{$\epsilon$-return mixing time of policy $\pi$} to be the
smallest value of $T$ such that $\pi$ guarantees an expected payoff of
at least $U(\pi) - \epsilon$; that is, it is the least $T$ such that
$U(s,\pi,t) \ge U(\pi) - \epsilon$ for all states $s$ and times $t \ge
T$.  
\edfn}
\shortv{
Formally, the \emph{$\epsilon$-return mixing time of an MDP $M$} is the  
least $T$ such that an optimal policy for $M$ guarantees an expected payoff
within $\epsilon$ of the optimal reward after running for at least time
$T$. (See \cite{KS02} for 
more details and motivation.)}
The following theorem 
shows that if the discovery probability is sufficiently low,
where ``sufficiently low'' means $D(1,t) < 1$ for all $t$ and $\sum_{t=1}^\infty D(1,t) <
\infty$, then the DM cannot learn to play near-optimally.
We define $\Psi(T) = \sum_{t=1}^T D(1,t)$.

\thm \label{t0}
If $D(1,t) <1$ for all $t$ and $\Psi(\infty) <\infty$, then there exists a
constant $c$ such that no algorithm can
obtain a reward that is guaranteed to be within $c$ of optimal for an MDPU
$M = (S,A,A_0,G,a_og_0,P,D,R)$,
even if $S$, $|A|$, and a bound on the optimal reward are known.
\ethm

Theorem \ref{t0} says that when $\Psi(\infty) < \infty$, it is impossible
for the DM to learn an optimal policy.
On the other hand, if $\Psi(\infty) = \infty$, then it is possible to
learn to play near-optimally.  
HRS present an algorithm
called URMAX, a variant of the RMAX algorithm \cite{BT02}, 
that learns near-optimal play.  
\thm \label{t1}
If $\Psi(\infty) =\infty$, then the URMAX algorithm computes a near-optimal policy.  \ethm

In fact, we can say even more.
If $\Psi(\infty) = \infty$, then the efficiency of the best
algorithm for determining near-optimal play depends on how quickly
$\Psi(\infty)$ diverges.   
HRS characterize the running time of
URMAX in terms of the function $\Psi$, and give lower bounds on the time 
required to learn near-optimally in terms of  $\Psi$.
These results show that URMAX learns to play near-optimally almost as
quickly as possible.  Specifically, it learns a policy with
an expected reward $\epsilon$-close to the optimal reward with probability 
$1-\delta$ in time polynomial in $|S|$, $|A|$, 
$1/\epsilon$, $1/\delta$, a bound $R_{\max}$ on the optimal reward, the $\epsilon$-return mixing time,
and the smallest 
$T$ such that $\Psi(T)\ge\ln(4N/\delta)$, whenever it is possible to do so.
The polynomial-time results are summarized in the next result.

\thm\label{cor:poly} It is possible to learn to play near-optimally in
polynomial time iff there exist constants $m_1$ and $m_2$ such that
$\Psi(T) \ge m_1 \ln(T) + m_2$ for all $T > 0$.
Moreover, if it is possible to learn to play near-optimally in
polynomial time, URMAX does so.
\ethm

\section{ANALYZING ROBOTIC PROBLEMS AS MDPUS} \label{sec:discrete}
As we said in the introduction, we apply the MDPU framework to robotic problems
such as having a humanoid robot learn to walk.
For such problems, we typically have a continuous
space of states and actions, where actions take place in continuous
time, and actions have a nontrivial duration.

Suppose that the original continuous problem can be characterized by a
continuous 
MDP $M_\infty$ (defined formally below).
We would like to find a ``good'' discretization $M$ of
$M_\infty$.  ``Good'' in this setting means that an optimal policy for
$M$ is $\epsilon$-optimal
for $M_\infty$, for some appropriate $\epsilon$.%
\footnote{A policy $\pi$ is $\epsilon$-optimal for an MDP $M$ if the
    expected average reward for a policy for $M$ is no more than
$\epsilon$ greater than 
the expected average reward of $\pi$.}
Clearly the level of
discretization matters.  
Too coarse a discretization results in an MDP whose optimal
policy is not $\epsilon$-optimal for $M_\infty$;  
on the other hand, too fine a
discretization results in the problem size becoming unmanageably large.  
For example, in order to turn a car on a smooth curve (without drifting),
the optimal policy is to slowly turn the steering wheel to the left and back,
in which the action varies smoothly over time. %
This can be simulated using a relatively coarse discretization of time. 
However, in order to make a sharp turn using 
advanced driving techniques like 
drifting, the steering wheel
needs to be turned at precise points in time, or else the car will go into an 
uncontrollable spin. In this case, a fine discretization in time is needed.

Unfortunately, it is often not clear what discretization level to
use in a specific problem. 
Part of the DM's problem is to find the ``right'' level of
discretization.  
Thus, we describe the problem in terms of a continuous
MDP $M_\infty$ and a sequence $((M_1,M_1'), (M_2,M_2'), \ldots)$, where $M_i$ 
is an MDPU with underlying MDP $M_i'$, for $i = 1, 2, \ldots$.  
Intuitively, $(M_1', M_2', M_3', \ldots)$ represents a sequence of finer
and finer approximations to $M_\infty$. 
\textbf{Continuous Time MDP with Continuous Actions over Time:}
To make this precise, we start by defining our model of continuous MDPs.
Let $M_\infty = (S_\infty, A_\infty, g_\infty,
P_\infty, R_\infty)$.  
$S_\infty$ is a continuous state space, which we identify with a compact
subset 
of $\IR^n$ for some integer $n >0$; that is, each state can be
represented by a vector $(s_1,\cdots,s_n)$ of real numbers.
For example, for a humanoid robot, the state space can be described by a vector which includes the robot's $(x,y,z)$ position, and the current positions of its movable joints.
\textbf{Actions:}
Describing $A_\infty$ requires a little care.
We assume that there is an underlying set of  
\emph{basic actions} $A_B$, which can be identified with a compact subset 
of $\IR^m$ for some $m >0$; that is, each basic action can be
represented by a vector $(a_1,\cdots,a_m)$ of real numbers.
For example, for a humanoid robot, the basic actions can be characterized by a tuple that contains the targeted positions for its movable joints.
However, we do not take $A_\infty$ to consist of basic actions.
\commentout{
We are interested in high-level actions such as grasping, drifting,
and walking. Such high-level actions usually correspond to a cluster of
basic action sequences, where each sequence is an instance of the action. 
For example, in order to grasp a mug, a robotic arm needs to perform a 
sequence of actions. There is a cluster of slightly different action sequences
that all achieve the purpose of the task, and are thus instances of the 
grasping action. Similarly, the $45^\circ$ drifting action refers 
to the cluster of action sequences that successfully drifts around a 
$45^\circ$ turn.
Thus, in this paper, we refer to high-level actions as clusters of
action instances; 
and actions as action instances, where each action instance is a sequence of 
basic actions over some time interval. 
}
Rather, an action is a \emph{path} of basic actions over time.
Formally, an action in $A_\infty$ is a piecewise continuous function 
from a domain of the form $(0,t]$ for some $t > 0$ to basic actions.  
Thus, there exist time points $t_0 < t_1 < \ldots < t_k$ with $t_0 = 0$
and $t_k = t$ such that $a$ is continuous in the interval
$(t_j,t_{j+1}]$ for all $j < k$. 
The number $t$ is the \emph{length} of the action $a$, denoted $|a|$.  
We use left-open right-closed intervals here; we think of the action in
the interval $(t_j, t_{j+1}]$ as describing 
what the DM does right
after time $t_j$ until time $t_{j+1}$.
By analogy with the finite case, $g_\infty(s)$ is the set of actions in
$A_\infty$ available at $s$.

\textbf{Reward and Transition Functions:}
We now define $R_\infty$ and $P_\infty$, 
the reward and transition functions.  In a discrete MDP, 
the transition function $P$ and reward function $R$ 
take as arguments a pair of states and an action.  Thus, for example,
$P(s_1,s_2,a)$ is the probability of transitioning from $s_1$ to $s_2$
using action $a$, and $R(s_1, s_2,a)$ is the reward the agent gets if a
transition from $s_1$ to $s_2$ is taken using action $a$.
In our setting, what matters is the path taken by a transition according
to $a$.  Thus, we take the arguments to $P_\infty$ and $R_\infty$ to be
tuples of the form $(s_1,s_c,a)$, where $s_1$ is a state, $a$ is an
action in $A_\infty$ of length $t$, and 
$s_c$ is a 
piecewise continuous function from $(0,t]$ 
to $S_\infty$.  Intuitively, $s_c$ describes
a possible path of states that the DM goes through when performing
action $a$, such that before $a$ starts, the DM was at $s_1$.%
\footnote{We are thus implicitly assuming that the result of performing
a piecewise continuous action must be a piecewise continuous 
state path.}  
Note that
we do not require that $\lim_{t\rightarrow 0^+} s_c(t) = s_1$.
Intuitively, this 
means that there can be a discrete change in state at the beginning
of an interval.
This allows us to capture the types of discrete changes considered in
\emph{semi-MDPs} \cite{Puterman1994}.  
We think of $R_\infty(s_1,s_c,a)$ as the reward for
transitioning from $s_1$
according to state path $s_c$ via action $a$. 
We assume that $R_\infty$ is bounded: specifically, 
there exists a  constant $c$ such that 
$R_\infty(s_1,s_c,a)< c\cdot |a|$.
\fullv{For state $s_1 \in S_\infty$ and action $a \in A_\infty$,} 
\shortv{For $s_1 \in S_\infty$ and $a \in A_\infty$,} 
we take $P_\infty(s_1,\cdot, a)$ to be a probability density function 
over state paths of length $|a|$ starting at $s_1$.  
$P_\infty$ is not defined for transitions starting at terminal 
states.
We require $R_\infty$ and $P_\infty$ to be continuous functions, 
so that if $(s_i,s_c^i,a_i)$ approaches $(s,s_c,a)$ (where all the state
sequences and actions have the same length $t$), then 
$R_\infty(s_i,s_c^i,a_i)$ approaches $R_\infty(s,s_c,a)$ and 
$P_\infty(s_i,s_c^i,a_i)$ approaches $P_\infty(s,s_c,a)$.  To make the
notion of ``approaches'' precise, we need to consider the distance
between state paths and the distance between actions.  Since we have
identified both states (resp., basic actions) with subsets of $\IR^n$ (resp.,
$\IR^m$), this is straightforward.
For definiteness, we define the distance between two vectors in $\IR^n$ using
the $L_1$ norm, so that $d(\vec{p},\vec{q})=\sum_1^n|p_i-q_i|$.
For actions $a$ and $a'$ in $A_\infty$ of the same length, 
define $d(a,a')=\int_{t=0}^{|a|}d(a(t),a'(t))dt$.
For state paths $s_c$ and $s_c'$ of the same length,
define $d(s_c,s_c')=\int_{t=0}^{|s_c|}d(s_c(t),s_c'(t))dt$.
Finally, define $d((s_c,a),(s_c',a'))=d(s_c,s_c')+d(a,a')$.
This definition of distance allows us to formalize the notion of
continuity for $R_\infty$ and $P_\infty$.  The key point of the continuity
assumption is that it allows us to work with discretizations, knowing
that they really do approximate the continuous MDP.

\textbf{Constraints on Actions:}
We typically do not want to take $A_\infty$ to consist of all possible
piecewise continuous functions.  
For one thing, some hardware and software restrictions will make 
certain functions infeasible. For example,
turning a steering wheel back and forth $10^{20}$ times in one
second can certainly be described by a continuous function, but is obviously
infeasible in practice. 
But we may want to impose further constraints on $A_\infty$ and $g_\infty(s)$.

In the discussion above, we did not place any constraints on the length
of actions.  
When we analyze problems of interest, there is typically an upper
bound on the length of actions of interest.  
For example, when playing table tennis using a robotic arm, 
the basic actions can be viewed as tuples, describing the direction of
movement of the racket, the rotation of the racket,
and the force being applied to the racket; actions are intuitively all
possible control sequences of racket movements that are feasible 
according to the robot's hardware and software constraints; 
this includes slight movements of the racket, strokes, and prefixes
of strokes.  
An example of a piecewise continuous action here would be to move the
racket forward with a fixed force
for some amount of time, 
and then to suddenly stop applying the force when the racket is close to
the ball. 
We can bound
the length of actions of interest to the time that a ball can be in the 
air between consecutive turns. 
\shortv{
We assume that, in any case, there is an upper bound $T$ on the length of all
actions.  This seems to be a reasonable assumption for all the
applications of interest to us here.}
\commentout{
Similarly, in grasping, the basic actions are tuples of angular velocities and 
linear velocities of the joints on the robotic arm, while actions 
are robotic arm movements
that correspond to high-level grasping actions, such as grasp, pinch,
clamp, or wrap \shortcite{LCZLZ}. Depending on the difficulty of the 
grasping object, an upper bound on the length of actions could be set;
for example, for grasping a mug, the upper bound could be set to 10 seconds.
Clearly, the exact choice of actions depends on the modeler.  We assume
that, in any case, there is an upper bound $T$ on the length of all
actions.  This seems to be a reasonable assumption for all the
applications of interest to us here.
}%

\commentout{
As usual, a policy in $M_\infty$ is a function from states to actions.
We are interested in optimal policies.  To make this precise, we need to
define the (average) reward of a policy.  This involves some technical
difficulties.  We avoid these difficulties by first considering
discretizations of $M_\infty$, defining the reward of a policy in the
discretized setting (where everything is finite), and then appealing to
continuity to compute the reward of a policy in $M_\infty$.  We do this
below, but first we bring awareness into the picture.
}

\textbf{Awareness:}
Even with the constraints discussed above, $A_\infty$ is typically
extremely large.  Of course,
not all actions in $A_\infty$ are ``useful''.
For instance, in the helicopter example, most actions
would crash the helicopter.
We thus consider \emph{potentially useful} actions.
(We sometimes call them just \emph{useful actions}.)
Informally, an action is potentially useful if 
 it is not \emph{useless}. A useless action is one that 
either destroys the robot, or leaves it in an uncontrollable state,
or does not change the state.
For example, when flying a helicopter, actions
that lead to a crash are useless, as are  actions 
that make the helicopter lose control.
More formally, given a state $s$, the 
set of useful actions at state $s$ are the actions that
transit to 
a different state in which the robot is neither destroyed nor uncontrollable.
Note that an action that crashes the helicopter in one state may not 
cause a crash in a different state. 
For robotics applications,
we say that a robot is \emph{aware} of an action 
if it identifies that action as a potentially useful action,
either because it has
been preprogrammed with the action (we are implicitly assuming that the robot
understands all actions with which it has been programmed)
or it has simulated the action.
\commentout{
Note that in order to check whether an action is potentially useful,
the robot has to have some knowledge of the action's transition function%
---perhaps not the complete transition function, but enough of it to be
able to tell whether the action is potentially useful.}
For example, a humanoid robot that has been pre-programmed
with only simple walking actions,
and has never tried running or simulated running before, 
would be unaware of running actions. 
Let $\bar{A}_\infty$ denote the useful actions in $A_\infty$, and 
let $\bar{A}_{\infty0}$ denote the useful actions that the robot
is initially aware of.  (These are usually the actions
that the robot has been pre-programmed with.)
\textbf{Discretization:}
We now consider the discretization of $M_\infty$.
We assume that, for each discretization level $i$, 
$S_\infty$ is discretized into a finite state space $S_i$ 
and $A_B$ is discretized into a finite basic action space $A_{Bi}$,
where $|S_1|\le |S_2|\le\ldots$
and $|A_{B1}| \le |A_{B2}| \le \ldots$.
We further assume that, for all $i$, there exists  $d_i >0$, 
with $d_i \rightarrow 0$,
such that for all states $s\in S_\infty$ and basic actions $a_B\in A_B$,
there exists a state $s' \in S_i$ and a basic action $a_{B'}\in A_{Bi}$  
such that $d(s,s')\le d_i$, and $d(a_B,a_B')\le d_i$.
Thus, we are assuming that the discretizations can give closer and
closer approximations to all states and basic actions.
At level $i$, we also discretize time 
into time slices of length $t_i$, where $T\ge
t_1>t_2>\ldots$.%
Thus, actions at discretization level $i$ are sequences of 
\emph{constant actions} of length $t_i$, where a constant action is a
constant function from $(0,t_i]$ to a single basic action.%
\footnote{Note that we are not assuming that the action space $A_{i+1}$
is a refinement 
of $A_i$ (which would further require $t_{i+1}$ to be a multiple of
$t_i$).} 
In other words, the action lengths at discretization level $i$ are
multiples of $t_i$.  
Thus, at discretization level $i$, there are $\sum_{l=1}^{\left\lfloor
T/t_i\right\rfloor} |A_{Bi}|^l$ possible actions. 
\fullv{
To see why, there are $|A_{Bi}|$ discrete actions at level $i$, and
action lengths  
must be multiples of $t_i$.
Thus, action lengths must have the form $lt_i$ for some $l \le \lfloor
 T/t_i\rfloor$.  There  
are $|A_{Bi}|^l$ actions of length
$l\times t_i$ at level $i$, and thus
$\sum_{l=1}^{\left\lfloor T/t_i\right\rfloor} |A_{Bi}|^l$
actions at level $i$. 
}%
Let $A_i'$ consist of this set of actions.
(Note that some actions in $A_i'$ may not be in $A_\infty$, since 
certain action sequences might be infeasible due to hardware and 
software constraints.)
Let $A_i\subseteq A_i'$ be the set of useful actions at level $i$.
\commentout{
There is a subtle difference in the meaning of discretizing the state space and discretizing the basic action space. To discretize the state space means we divide the continuous state space into a number of  smaller continuous state spaces, where  each smaller state space represents a state (see example below).
On the other hand, to discretize the basic action space means to choose values from each dimension of the basic action space, such that all basic actions at the level must be tuples composed of the corresponding chosen values.
For example, for a problem with one dimension in both its state space
and its basic action space. Call the dimension as $x$. Suppose $x\in
[0,1]$. In the state space, discretizing $x$ into 2 values means
dividing its domain into 2 smaller domains; for example, $[0,0.5)$ and
  $[0.5,1]$. Given a state $(x)$, if $x\in [0,0.5)$, then the DM is in
    state 1; and if $x\in [0.5,1]$, the DM is in state 2. 
On the other hand, in the basic action space, discretizing $x$ into 2
values means to choose 2 values from $[0,1]$, for example, suppose we
choose $0.25$ and $0.75$. Then, the discretization has only two basic
actions $(x)=(0.25)$, and $(x)=(0.75)$. 
So discretizing the states means multiple states in the continuous
state space may map to one discrete state, while discretizing the
basic actions means there are only a finite number of basic actions
you can choose from. 
  }
  Let $M_i$ be the MDPU %
where $S_i$ and $A_i$ are defined above;
$A_{\infty 0}$ is the  
set of useful actions that the DM is initially aware of;
$g(s)$ is the set of useful actions at state $s$; $g_0(s)$ is the set 
of useful actions that the DM is aware of at state $s$;
and
the reward function $R_i$ is just the restriction of $R_\infty$ to $A_i$ 
and $S_i$. For $s_1\in S_i$ and $a\in A_i$, we take $P_i(s_1,\cdot,a)$
to be a probability distribution over $Q_i^{|a|}$, the set of 
state paths of length $|a|$ that are piecewise constant and each
constant section has a length that is a multiple of $t_i$.
For a state path $s_c\in Q_i^{|a|}$, let $P_i(s_1,s_c,a)$ 
be the normalized probability of traversing a state sequence
that  
is within distance $d_i$ of state sequence $s_c$ when playing action
$a$ starting from state $s_1$. 
Formally, $P_i(s_1,s_c,a) = 
(\int_{\{s_c':d(s_c,s_c')\le d_i\}} dP_\infty(s_1,\cdot,a))/c$,
where $c = \sum_{s_c \in Q_i^{|a|}} \int_{\{s_c':d(s_c,s_c')\le d_i\}}
dP_\infty(s_1,\cdot,a)$ is a normalization constant. 
Since the robot is not assumed to know all useful actions at any specific discretization
level, it needs to explore for the useful actions it wasn't aware of. 
Finally, given a specific exploration strategy, $D_i(j,t)$ describes the 
probability of discovering a new useful action at discretization level $i$, 
given that there are $j$ undiscovered useful actions at level $i$, 
and the robot has explored $t$ times without finding a useful action.
We model exploration using $a_0$; every time the robot explores, it is
playing $a_0$.

It remains to define the discretization of an action in $A_0$.  
In order to do this, for $a\in A_\infty$, define $a_i\in A_i$
to be \emph{a best  
approximation to $a$ in level $i$} if $|a_i|$ is the 
largest multiple of $t_i$ that is less than or equal to $|a|$, 
and $\int_0^{(|a_i|)} d(a(t),a_i(t)) dt$ is minimal among 
actions $a' \in A$ of length $|a_i|$.  
Intuitively, $a_i$ is an action in $A_i$ whose length is as close as
possible to that of $a$ and, among actions of that length, 
is closest in distance to $a$. The action $a_i$ is not unique.
For $a\in A_0$, define its discretization at level $i$ to be 
a best approximation to $a$ at that level. 
When there are several best approximations, we choose any one of them. 
\textbf{Policies:}
As usual,
a policy $\pi$ in $M_i$ is a function from $S_i$ to $A_i$. 
We want to compute $U_{M_i}(s,\pi,t)$, 
the expected average reward over time $t$ of $\pi$ started in 
state $s\in S_i$.  
Let $a_j\in A_i$ and $s_{cj}$ be a state sequence in $Q_i^{|a_j|}$,
for $j = 0, \ldots, l$.
Say that a sequence $((a_0,s_{c0}),(a_1,s_{c1}),\cdots,(a_l,s_{cl}))$ %
is \emph{a path compatible with policy $\pi$ starting at $s$} if  
$\pi(s)=a_0$ 
and $\pi(s_{cj}(|a_j|))=a_{j+1}$ for all
$0\le j\le l-1$.  Let $I_{s,t}^{\pi}$ consist of all paths 
$((a_0,s_{c0}),(a_1,s_{c1}),\cdots,(a_l,s_{cl}))$ starting at $s$
compatible with $\pi$ such that 
	$\sum_{j=0}^l|a_j|\le t<\sum_{j=0}^{l+1}|a_j|$, where
$a_{l+1}=\pi(s_{cl}(|a_l|))$. 
Essentially, when computing $U_{M_i}(s,\pi,t)$, we consider the
expected reward over all maximally long paths that have 
total length at most $t$.
Thus, $U_{M_i}(s,\pi,t)=\sum_{p\in I^{\pi}_{s,t}}
P^*_i(p)\frac{R^*_i(p)}{t}$, 
where,
given a path $p=((a_0,s_{c0}),(a_1,s_{c1}),\cdots,(a_l,s_{cl}))$,
$P_i^*(p)=\Pi_{j=0}^{l} P_i(s_{cj}(0),s_{cj},a_j)$, and
$R_i^*(p)=\sum_{j=0}^{l} {R_i(s_{cj}(0),s_{cj},a_j)}$.
Now that we have defined the average reward of a policy at
discretization level $i$, we can define the average reward of a policy
in $M_\infty$.  
Given a discretization level $i$, let $\pi_i$ be a \emph{projection}  of
$\pi_\infty$ at level $i$, defined as follows:
for each $s_i\in S_i$, define $\pi_i(s_i)$ to be an action $a_i \in
A_i$ such that $a_i$ is a best approximation to $\pi(s_i)$ at level $i$,
as defined above.
As mentioned, there might be several best approximations; $a_i$
is not unique.
Thus, the projection is not unique.  Nevertheless, 
we define $U_{M_\infty}(s,\pi_\infty,t)$ to be 
$\lim_{i\rightarrow\infty}U_{M_i}(s,\pi_i,t)$, where $\pi_i$ is a
projection of $\pi$ to discretization level $i$.
The continuity of the transition and reward functions guarantees that
the limit exists and is independent of the choice of projections.

We now consider how the URMAX algorithm of Section~\ref{sec:mdpu} can be
applied to learn near-optimal policies.  
We use URMAX at each discretization level.  Note that URMAX never
terminates; however, it eventually learns to play near-optimally
(although we may not know exactly when).  The time it takes to learn to
play near-optimally depends on the exploration strategy.
The next theorem consider brute-force searching, where,
at discretization level $i$, 
at each discretization level $i$, all actions in $A_i'$ are exhaustively
examined to find useful actions.
(The proof of this and all other theorems can be found in the
supplementary material.)
\thm \label{t11}
Using brute-force exploration, 
given $\alpha>0$ and $0<\delta<1$, we can find an 
$\alpha$-optimal policy in $M_{\infty}$ with probability at least
$1-\delta$ 
in time polynomial in $l$, $|A_l'|$, $|S_l|$, $1/\alpha$, $1/\delta$,
$R_{\max}^l$, and $T^l$,  
where $l$ is the least $i$ such that the optimal policy for $M_i'$ is
$(\alpha/2)$-optimal 
for $M_\infty$, 
$R_{\max}^l$ is the maximum reward that can be obtained by a transition in $M_l'$,
and $T^l$ is the $\epsilon$-return mixing time for $M_l'$.
\ethm

Although brute-force exploration always learns a near-optimal
policy, the method can be very inefficient, since
it exhaustively checks all possible actions to find the useful ones.  
Thus, at discretization level $i$, it needs to check $\sum_{l=1}^{\left\lfloor T/t_i\right\rfloor}|A_{Bi}|^l$ actions,
and as $i$ grows, the method soon becomes impractical. On the other hand,
the result is of some interest, since it shows that even when there are
infinitely many 
possible levels of discretizations, a method as simple as brute-force 
exploration suffices.

When the number of possible actions is huge, the probability of finding a
potentially useful action can be very low. In this case, making use of 
an expert's knowledge or imitating a teacher's demonstration can often greatly
increase the probability of finding a useful action.
We abstract the presence of an expert or a teacher by assuming that
there is some constant $\beta > 0$ such that $D(1,t) \ge \beta$ for
all $t$.  Intuitively, the presence of a teacher or an expert guarantees
that there is a minimal probability $\beta$ such that, if there is a
new action to be found at all, then the probability of finding it is at
least $\beta$, no matter how many earlier failed attempts there have
been at finding a useful action.
\fullv{
For example, Abbeel and Ng (2005) study the problem of robotic
helicopter flying. 
They assume that they have a teacher that will help demonstrate how to fly.
Their assumptions imply that there is a constant $\beta > 0$
such that $D(1,t) \ge \beta$.%
\footnote{Specifically, if we take a flight with reward $\epsilon$-close
to the flight demonstrated by the teacher to be a useful
action, and take $a_0$ be the process of 
performing $h$ iterations of the main loop in their algorithm, where
$h=
\frac{64HR_{\max}}{\epsilon}(2+c\log\frac{64H^2R_{\max}|S|^3|A|}{\epsilon})$,
then the probability 
of finding a useful action is at least 
$1-e^{\frac{\epsilon}{(1+c)32HR_{\max}}}$, where 
$c=\frac{16^2H^2R_{\max}^2|S|^3|A|}{4\epsilon^2}$;
$H$ is the horizon, so that the procedure must terminate after
$H$ steps; 
$R_{\max}$ is the maximum reward;
$|S|$ is the number of states; and $|A|$ is the number of actions.}
}%
Using apprentice learning lets us improve Theorem~\ref{t11} by replacing
the $|A'_l|$ component of the running time by $|A_l|$; thus, with
apprentice learning, the running time depends only on the number
of useful actions, not the total number of potential actions.  The
savings can be huge.
\thm \label{t12}
Using an exploration method where $D_i(1,t)\ge \beta$ for all $i,t>0$
(where $\beta\in(0,1)$ is a constant),  
for all $\alpha>0$ and $0<\delta<1$, we can find an 
$\alpha$-optimal policy in $M_{\infty}$ with probability at least $1-\delta$
in time polynomial in $l$, $|A_l|$, $|S_l|$, $1/\beta$, $1/\alpha$,
$1/\delta$, $R_{\max}$, and $T^l$,  
where $l$ is the smallest $i$ such that the optimal policy for $M_i'$ is
$(\alpha/2)$-optimal 
to $M_\infty$, 
$R_{\max}^l$ is the maximum reward that can be obtained by a transition in $M_l'$, 
 and $T^l$ is the $\epsilon$-return mixing time for $M_l'$.
\ethm

\section{HUMANOID ROBOT WALKING}
\begin{figure}
\centering
\subfigure{
\includegraphics[width=0.65\linewidth]{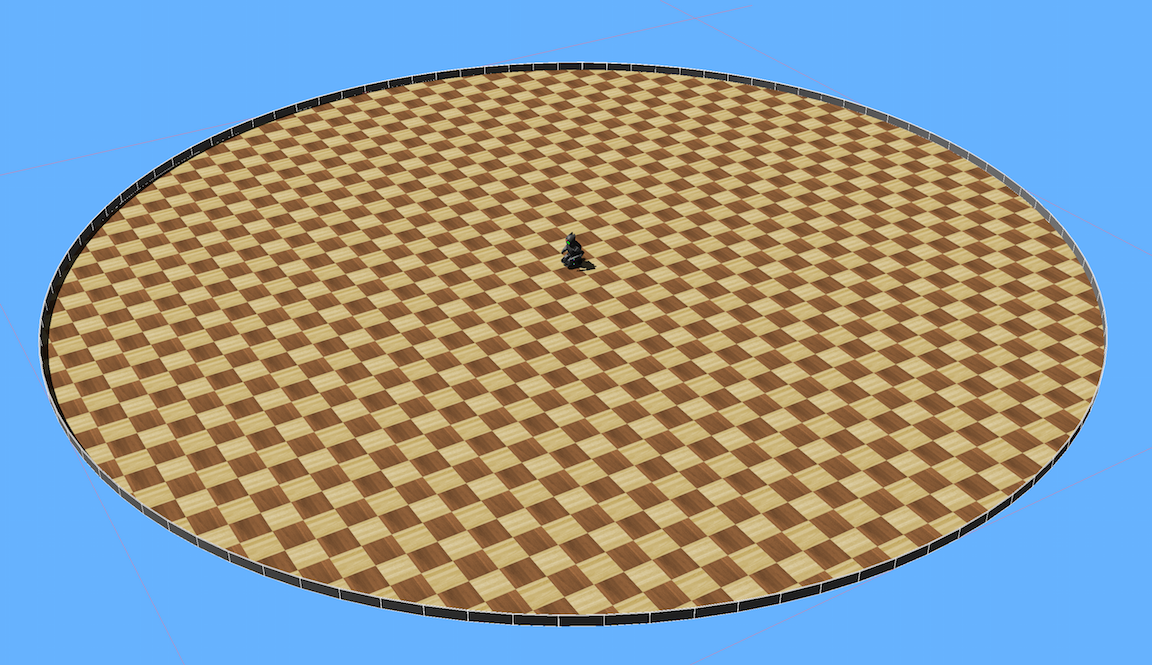}
\includegraphics[width=0.2\linewidth]{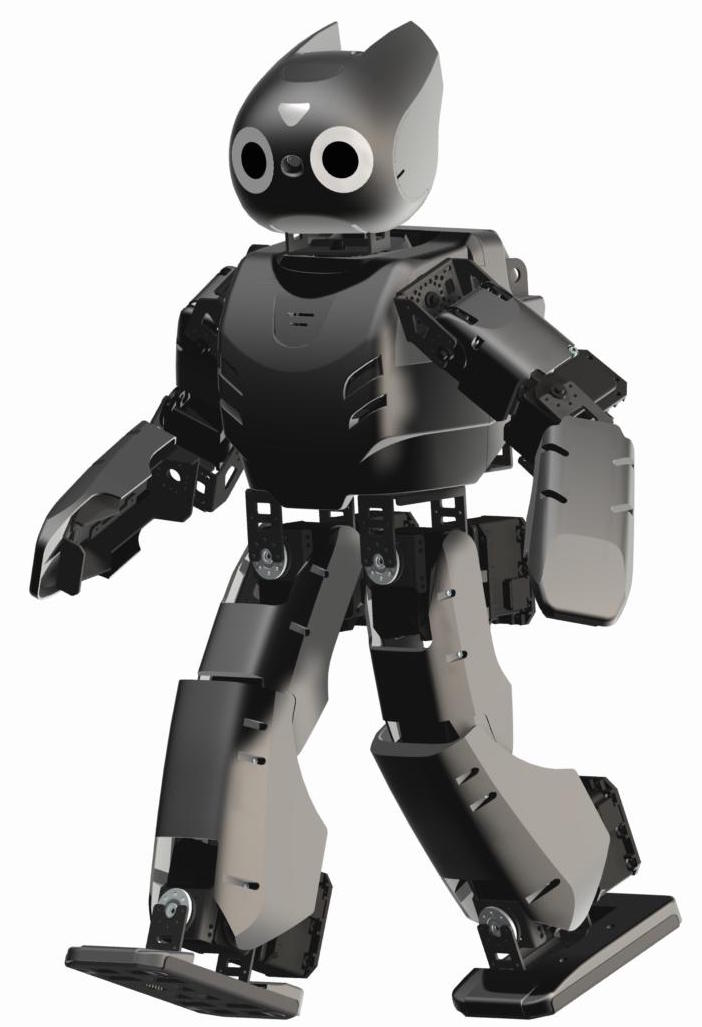} 
}
\caption{The arena with the robot at the center; and the robot.}\label{arena}
\end{figure}

We consider the problem of a humanoid robot with 20 joint
motors (which we sometimes call just ``joints'') learning to walk on its own.
More precisely, we require the robot to move from the center of an
arena to its  boundary; we take any reasonable motion to be
``walking''.  (Figure~\ref{arena} shows the robot and
the arena in which it must walk.)

\subsection{The continuous MDP}
We start by defining a continuous $M_\infty$ for the robot problem.
A state $s\in S_\infty$ is of the form $s=(w_1,\cdots,w_{23})\in
\IR^{23}$, where $(w_1, w_2, w_3)$ give the position of the robot's
center of mass 
and $(w_4,\cdots,w_{23})$ are the current positions of the robot's 20
joint motors. We define the domain of each dimension as
follows: since the radius of the arena is 5 meters,  $w_1,w_2\in
[-5,5]$; since the robot's height is 0.454 meters, $w_3\in
[0,0.4]$ (we do not expect the robot's center of mass to be
higher than 0.4). Each joint motor has its specific range of mobility,
which determines the domain of the corresponding dimension. For example,
$w_5\in [-3.14,2.85]$ represents the current position of the robot's
left shoulder. The mobility range for all joint motors are intervals in
$[-\pi,\pi]$. 

The basic actions $a\in A_B$ are of the form
$a=(v_1,\cdots,v_{20})\in \IR^{20}$, where $v_i$ is the target
position for the robot's $i$th joint motor. The domain of each
dimension is the mobility range for the corresponding joint motor. For
example, $v_2$, which corresponds to the left shoulder, has mobility
range $[-3.14,2.85]$; $v_2=2.85$ means to move the robot's left
shoulder forward as far as possible.
Since walking is composed of repeated short movements that are
typically no longer than half a second, we set $T=0.512$ seconds. 
Thus, $A_\infty$, the set of useful actions, consists of piecewise continuous
functions that map from time to basic actions and comply with
the robot's hardware and software limitations, 
of length $t < 0.512$ seconds.

We now define $R_\infty$ and $P_\infty$.
Intuitively, the robot obtains a reward for gaining distance from the
center of the arena.  
If the coordinates of the center of the arena are given by $s_0 =
(s_0[1],s_0[2])$, then 
$R_\infty(s_1,s_c,a)=dis(s_0,s_c(|a|))-dis(s_0,s_1)$, where
$dis(s_0,s_1)=\sqrt{(s_0[1]-s_1[1])^2+(s_0[2]-s_1[2])^2}$ is the
$L_2$-norm distance between $s_0$ and $s_1$ on the $(x,y)$-plane. The
reward 
could be negative, for example, if the robot moves back towards the
center of the arena. 

By definition, $P_\infty(s_1,\cdot,a)$ is a probability distribution over
state sequences of length $|a|$ starting at $s_1$. For example, if
the robot slowly moves its right leg forward while staying balanced,
the state path taken by the robot is a deterministic path. On the
other hand, if $a$ is the action of turning around quickly, 
$P_\infty(s,\cdot,a)$ is distribution over various ways of falling down.

\subsection{Discretizations}\label{subsec:discretization}
We now define $M_i$ and $M_i'$.
In our experiments we considered only levels 2 and 3
(level 1 is uninteresting since it has just one state and one action),
so these are the
only levels that we describe in detail here.  (These turn out to suffice
to get interesting walking behaviors.)  At these levels, we
discretized more finely the joints corresponding to the left and right
upper and lower leg joints and the left and right ankle joints, since
these turn out to be more critical for walking.
(These are components $(w_{14},\cdots,w_{19})$ in the state
tuples and $(v_{11},\cdots,v_{16})$ in basic-actions tuples.)  We
call these the \emph{relevant dimensions}.  We assume that the six
relevant state and actions components have $i$ possible values at level
$i$, for $i = 2,3$, as does $w_3$, since this describes how high off
the ground the robot is (and thus, whether or not it has fallen).
All other dimensions take just one value.
We took $t_2=t_3$ to be $128$ms.  
Since $T=0.512$s, 
an action contains at most $\left\lfloor
T/t_i\right\rfloor=4$ basic actions.

$A_{\infty 0}$ is the set of preprogrammed actions. 
We preprogram the robot with a simple sitting action that lets the
robot slowly return to its initial sitting gesture. When
we consider apprenticeship learning, we also assume that the robot is
preprogrammed with a ``stand-up'' action, that enables it to stand up
from its initial sitting position.  (Intuitively, we are assuming that
the expert taught the robot how to stand up, since this is useful
after it has fallen.)

$A_i'$ is the set of potential actions at level $i$.
Given our assumptions, for $i = 2,3$, at level $i$,
there are $(i^6)^4$ potential actions (there are $i$ possible values
for each of the six relevant dimensions, and each action is a sequence
of four basic actions).
Thus, at level 3, there are
$(3^6)^4=$282,429,536,481 potential actions.  
As we mentioned, a useful action is an action that moves the
robot without making it lose control. Here, an action is useful if it
moves the robot without resulting in the robot falling down.
At both levels 2 and 3, more than 80 useful actions were found in our
experiments.  The most 
efficient action found at level 3 was one  where the right leg moves
backwards, immediately followed by the left leg, in such a way that
the robot maintains its balance
at all times. By way of contrast, turning the body quickly makes the
robot lose control and fall down, so is useless. 

For $s_1\in S_i$,
$a\in A_i$, and $s_c\in Q_i^{|a|}$, $P_i(s_1,s_c,a)$ is the normalized probability of traversing a state sequence that is $d_i$ close to $s_c$, a sequence of states in $S_i$, where we define $d_i=\frac{12\pi}{i}+28\pi+20.4$. 
So $d_i$ decreases in $i$, and discretizations at a higher
level better approximate  the continuous problem.  
All basic actions in $A_B$ are within distance $d_i$ of a basic action
in $A_{Bi}$ and all states in $S$ are within $d_i$ of a state in
$S_i$. 
Let $s\in S$, and let $s_i$ be the closest state to $s$ in $S_i$.
It is easy to check that $d(s,s_i)\le d_i$ for $i = 2,3$.

The $D_i$ function depends on the exploration method used to discover new actions. In our experiment, we used two exploration methods: brute-force exploration and apprenticeship-learning exploration.

At discretization level $i$,
using brute-force exploration, we have 
$D_i(|A_i|,t)=\frac{|A_i|}{|A_i'|}$, 
since there are $|A_i|$ useful actions and $|A_i'|$ potential
actions, and we test an action at random.
With apprenticeship learning, 
we used following hints from a human expert to increase the
probability of discovering new actions:  
(a) a sequence of 
moving directions that, according to the human
expert, resembles human walking;%
\footnote{The sequence gives directions only for certain joints, without
specific target values, leaving the movement remaining  joints open
for experimentation.}
(b) a preprogrammed stand-up action;
(c) the information that an action that is symmetric to a useful
action is also likely to be useful (two 
actions are symmetric if they are exactly the same except that the
target values for the left joints and those for the right joints are
switched). 
We also use a different discretization: the ankle joint was
discretized into 10 values.
The human expert suggests more values in the ankle joints 
because whether or not  the robot falls depends critically on the
exact ankle joint position.
These hints were provided before the policy starts running; the
discretization levels are set then too.  There were no  further
human-robot interactions.

\subsection{Experiments}
For our experiments, we simulated  
DARwIn OP, a commercially available humanoid robot.
The simulations were conducted on Webots PRO platform 8.2.1 using a MacBook Pro with 2.8GHz Intel Core i7 Processor, 16GB 1600 MHz DDR3 memory, 0.5TB Flash Storage Mac HD, on OS X Yosemite 10.10.5. 
We modeled the robot walking problem as an MDPU, and implemented the URMAX algorithm 
\fullv{to solve the problem } 
using programming language Python 2.7.7.

As we said, given the number of actions involved, we conducted
experiments only for discretization level 2 and 3.
Both sufficed to enable the robot to learn to walk, using a generous
notion of ``walk''---more precisely, they sufficed to enable the robot
to learn to locomote to the boundary of the arena.
As mentioned, two exploration methods were used:  brute-force
exploration and apprenticeship-learning exploration.
One trial was run for brute-force exploration at each of levels 2 and
3, and one trial was run for apprenticeship learning
at level 2. Each trial took 24 hours. More than 15 stable gaits were
found in total, where a gait is \emph{stable} if it enables the robot
to move from the center of the arena to the boundary without falling.
In addition, more 
than 400 useful actions were found. The best gait among all stable
gaits achieved a velocity of 0.084m/s,
which seems reasonable, given that the best known walking speed of DARwIn-OP is
0.341m/s \cite{BWFM2013}. 
Given more time to experiment, we would expect the performance to
improve further.

\begin{table}[]
\scriptsize
\centering
\label{table:performance}
\begin{tabular}{|l|l|l|l|}
\hline
& \begin{tabular}[c]{@{}l@{}}Brute-force\\ (level 2)\end{tabular} & \begin{tabular}[c]{@{}l@{}}Brute-force\\ (level 3)\end{tabular} & \begin{tabular}[c]{@{}l@{}}Apprenticeship\\ learning (level 2)\end{tabular} \\
\hline
$|S_i|$ & 130  & 1460 & 3200 \\
\hline
$|A_{Bi}|$  & 64  & 729 & 1600 \\
\hline
$|A_i|$  & 16777216 & 282429536481 & 6553600000000 \\
\hline
$t_i$ (ms) & 124 & 124 & 124 \\
\hline
\begin{tabular}[c]{@{}l@{}}Length of\\action (ms)\end{tabular}& 496   & 496   & 496 \\
\hline
\begin{tabular}[c]{@{}l@{}}Execution\\ time (hours)\end{tabular} & 24 & 24 & 24 \\
\hline
\begin{tabular}[c]{@{}l@{}}Best avg rwd\\   (m/action)\end{tabular}& 0.043486  & 0.067599  & 0.083711\\
\hline
\begin{tabular}[c]{@{}l@{}}Num of useful\\ actions found\end{tabular} & 131 & 89 & 180 \\
\hline
\end{tabular}
\caption{Performance comparisons.}
\end{table}
\normalsize

\begin{figure}
\centering
\subfigure{
\includegraphics[width=0.25\linewidth]{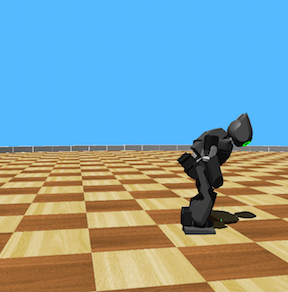}
\includegraphics[width=0.25\linewidth]{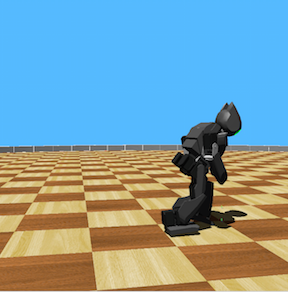} 
\includegraphics[width=0.25\linewidth]{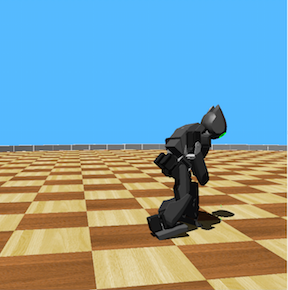} 
\includegraphics[width=0.25\linewidth]{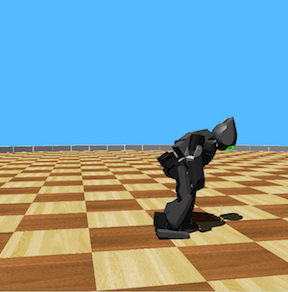} 
}
\caption{A backward gait (from left to right).}\label{back gait}
\end{figure}

\begin{figure}
\centering
\subfigure{
\includegraphics[width=0.25\linewidth]{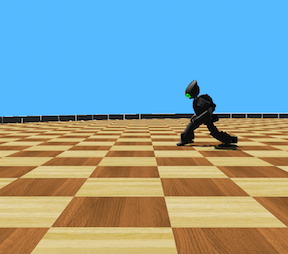}
\includegraphics[width=0.25\linewidth]{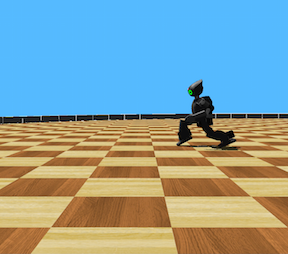} 
\includegraphics[width=0.25\linewidth]{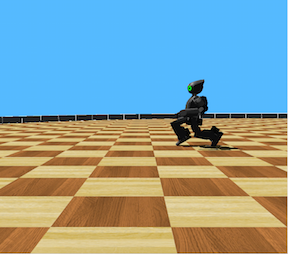} 
\includegraphics[width=0.25\linewidth]{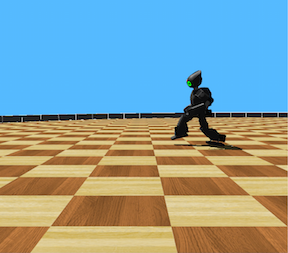} 
}
\caption{A forward gait (from left to right).}\label{forward gait}
\end{figure}
The robot successfully learned gaits of various
styles, including both forward and backward gaits (see Figures
\ref{back gait} and \ref{forward gait}), both efficient and
inefficient gaits, gaits that resemble human walking and
the ones that do not.
Somewhat surprisingly, the best gait actually walks
backwards. (Videos of some of the gaits 
and a demo of the learning process 
can be found at
https://youtu.be/qW51iInpdV0.)
As shown in Table 1, 
as the discretization level increases,  
both the velocity of the best gait and the number of useful actions
found increase.  
This agrees with the expectation that finer discretization better
approximates the continuous problem, and thus gets an expected reward
closer to the optimal reward of the continuous problem. 
Apprenticeship learning resulted in more useful actions
than the brute-force exploration and in gaits with a higher average
reward. Again, this is hardly surprising; the hints provided by
the human expert increases the probability of finding useful actions.  
On the other hand, when the expert gives ``bad'' hints,
the robot performs worse than with brute-force exploration. 

With regard to comparisons under the same setting, we implemented
two baseline comparisons 
for bipedal walking.
The first tries random sequences of 
actions; the second searches for useful actions and then repeats each 
useful action in an attempt to find stable gaits. (The motivation for 
repeating actions is that human-like walking is in fact a 
rhythmic/cyclic movement of a repeated action.)  Using the same
setting as our experiments with URMAX,
the first baseline found no stable gait in 24 hours, and the maximum 
distance the robot traveled before falling down was 0.342 meters; the 
second baseline found one stable gait in 24 hours, and the maximum 
distance traveled was 5 meters (the maximum distance possible from the 
center of the arena to the boundary) with a speed of 0.0058 m/s.
Our approach found several stable gaits at each discretization
level and the  
distance traveled using each stable gait is 5 meters with a maximum speed 
of 0.083m/s (10 times better than the second baseline).
Our approach, using MDPUs, requires no knowledge on the kinematics of the robot
other than the number of joints and the moving range of each joint.
Moreover, it makes no assumptions about the moving pattern of the
resulting gait; 
for example, we do not assume that a gait must be cyclic, or
symmetric between left and right joints, nor do we specify the
length of a gait.  Although we do specify the length of a useful action,
a gait could be composed of a single or multiple useful
actions.  Given the few assumptions and little prior knowledge assumed, 
the performance of the robot seems quite reasonable. 
More importantly, the experiment proves
that the use of MDPUs enables the robot to learn
useful new maneuvers  (walking, in this case) by itself, with minimum
human input.
\commentout{
\textbf{Future work}
In the future, we plan to apply our framework to more robotic
tasks, such as learning to run and to walk up and down 
stairs. We are also interested
in seeing of the robot can simulate learning in a baby, for example,
by limiting the robot's abilities initially so it must crawl before it walks.
}

\section{RELATED WORK} \label{sec:related}
There has been work on optimal policy learning in MDPs using computational resources. 
Kearns and Singh's \citeyear{KS02} $E^3$ algorithm 
guarantees polynomial bounds on the resources required to achieve
near-optimal  
return in general MDPs;
variants and extensions of this work can be found in \cite{BT02,KKL03,KK99}.
However, algorithms such as $E^3$ usually require 
the exploration of the entire MDP state/action space.
This becomes impractical in our setting, where the number of actions is
extremely large. 
In such cases, several exploration methods have been employed to help
find  useful actions.  For example, Abbeel and Ng
\citeyear{AN05ICML} utilize 
a teacher 
demonstration of the desired task to guide the exploration;
Dearden et al. \citeyear{DFA99} utilize the value of information to
determine the sequence of  
exploration. %
\fullv{ Guestrin et al. 
\citeyear{GPS2002} make use of approximate linear
programming, and focus on exploring states that 
directly affect the results of the planner; Kakade et al. \citeyear{KKL03} proved
that in certain situations, the amount of time required to compute a
near-optimal policy depends on the \emph{covering number} of the state space,
where, informally, the covering number is the number of neighborhoods required
for accurate local modeling; 
}%
Other papers (e.g., \shortcite{DKG98,NMKD98}) consider
MDPs with large action spaces.

We are far from the first to consider MDPs with continuous time. 
For example, 
\emph{semi-MDPs} (SMPDs) and \emph{continuous-time MDPs} have continuous
time \cite{Puterman1994}.
However, these models have discrete actions that can be taken
instantaneously, and do not consider continuous actions taken 
over some duration of time. 
In \emph{Markov decision drift processes} \cite{HV84}, the state does
not have to stay constant between successive actions (unlike an SMDP),
and can evolve in a deterministic way according to what is called a
\emph{drift function}. But Markov decision drift processes do not have
actions with probabilistic 
outcomes that take place over an interval of time. 
\shortv{They}
\fullv{Hordijk and van der Duyn Schouten \citeyear{HV84}} 
make significant use
of discrete
approximations to compute optimal policies, just as we do. 
There has also been work on MDPs with continuous state space and 
action space 
(e.g., \shortcite{AM07,FDMW04}), but with discrete time.
For our applications, we need time, space, and actions to be continuous;
this adds new complications.
In control theory, there are methods for controlling continuous 
time systems where system transitions are linear function of 
state and time \cite{Zin89}.  These can be extended to non-linear 
systems \cite{Khalil02}. 
However,
the transitions in these systems are usually deterministic, and 
they do not deal with rewards or policies.
Sutton, Precup, and Singh \citeyear{SPS99} consider high-level actions
(which they call \emph{options})  
that are taken over a duration of time (such as ``opening a door''), 
but they view time as discrete, which
significantly simplifies the model.
Rachelson, Garcia, and Fabiani \citeyear{RGF08} consider
continuous actions over some time interval, however, they assume there
are decision epochs, which are the only time points where rewards are
considered. In our model, the rewards depend on the entire
state sequence that the system traverses through while an action is taken.
While this makes the model more complicated, it 
seems more appropriate for the problems of interest to us.
There has also been a great deal of work on bipedal robot walking,
since it is a fundamental motor task for which
biological systems significantly outperform current robotic systems
\cite{TZS2005}.
There have been three main approaches for solving the task: 
\begin{itemize}
\item The
first approach describes the kinematics of the robot in detail using
non-linear and linear equation systems, then solves these systems to
obtain desirable trajectories.
See, for example,
\cite{HYKKAKT2001,GZ2006,KKKFHYH2003,KPO2007,SSC2010,XCLN2012}.  
\item
The second approach uses genetic algorithms
\cite{CL1997,HAF2000,PGLRT2009}. 
The traits describing a gait are taken to be 
the genes in a genetic algorithm.
Different gaits (i.e., settings of the parameters) are evaluated in
terms of features such as stability and velocity;
The most successful gaits are retained, and used to produce the next
generation of gaits through selection, mutation, inversion, and
crossover of their genes. 
\fullv{
This approach can also be used for to learn quadrupedal and
nine-legged walking. See, for example,
\cite{CV2004,PT2011,PTC2011,ZBL2004}. }
\item The third approach uses gradient learning, which starts with
either a working gait or a randomly initialized gait. It then improves
the gait's performance by changing its parameters,  using
machine-learning methods (such as neural networks) to find the most
profitable set of changes in the parameters.
See, for example,
\cite{BWFM2013,KU2003,SLMJA2015,TZS2005}.
\fullv{
this approach is also used in quadrupedal walking \cite{KS2004}. %
}
\end{itemize}

Since the first approach requires a full description of the robot's
kinematics, as well as composing and solving  a non-linear
system, it requires a great deal of human input. Moreover, its application is
limited to walking problems.
The approach is unable to produce gaits other
than those specified by human (e.g., to walk forward by stepping
forward the left and the right legs in turn under a specific speed). 
Both the second and the third approach require little human input
(when starting from random gaits), and may produce a variety of
gaits. Both also have the potential to be generalized to problems other
than bipedal walking. However, both are heuristic search algorithms,
and have no theoretical guarantee on  their performance. In contrast,
our method produces a variety of gaits,
provides a general framework for solving robotic problems, and
produces a 
near-optimal policy in the limit. Moreover, our method requires minimum
human input, although, as the experiments show, it does better with
more human input.

A comparison of our method and the genetic algorithm may provide
insights into both methods. Although the two approaches seem
different, the 
searching process made possible by selection, mutation, inversion, and
crossover in a genetic algorithm can be viewed as a special case of
the \emph{explore} action
in an MDPU.
Conversely, the \emph{explore} action in an MDPU for the robot can be
roughly viewed as 
searching for a set of genes of unknown length (since a gait can be
understood as a continuous action over an uncertain amount of time, 
composed of one or more shorter actions, where each shorter
action is described by a set of parameters).  Our approach can be
viewed as being more flexible than a genetic algorithm; in a genetic
algorithm, the length of the chromosome (i.e., the number of
parameters that describe the gait) is fixed; only their values
that give the best performance are unknown.  

The recent work of 
Mordatch et al \citeyear{MMEA2016} also provides a
general approach for reinforcement learning in robot tasks. Like
us, they require prior knowledge only of the mobility range of each of the
robot's joints, and not their kinematics;  they
also model the problem as an MDP. Their goal is to find an optimal
trajectory (i.e., a sequence of states), such that when followed,
performs a desired task (such as reaching out the robot's hand to a
desired position) with minimal cost. They use neural networks to solve
the cost-minimization problem.
Thus, their approach does not have any guarantees of (approximate) optimality
of the performance. 
Moreover,
the complexity of their approach grows quickly as the length of the
trajectory grows (while ours is polynomial  in the number of useful
actions, states visited, and the difficulty of discovering new
actions, and thus is not significantly affected by the length of the
trajectory). 
That said, Mordatch et al.'s method has successfully learned a
few relatively simple tasks on a physical DARwIn OP2 robot, including
hand reaching and leaning the robot's torso to a desired position
\cite{MMEA2016}, although it has not yet been applied to walking.

\section{CONCLUSION}
We have provided a general approach that allows robots to
learn new tasks on their own. We
make no assumptions on the structure of the tasks to be learned. 
We proved that in the limit, the method gives a near-optimal policy. 
The approach can be easily applied to
various robotic tasks.  We illustrated this by applying it to 
the problem of bipedal walking.  Using the approach, a
humanoid robot, DARwIn OP, was able to learn various walking gaits via simulations
(see https://youtu.be/qW51iInpdV0 for a video).
We plan to apply our approach to more robotic
tasks, such as learning to run and to walk up and down 
stairs.
We believe the process will be quite instructive in
terms of adding useful learning heuristics to our approach, both
specific to these  tasks and to more general robotic tasks.
We are also interested
in having the robot simulate learning to walk in the same way a baby
does, for example,
by limiting the robot's abilities initially, so that it must crawl before it
walks.
Part of our interest lies in seeing if such initial limitations actually make
learning more efficient.
\commentout{
There has been a lot of work on grasping (see, for example \cite{HL06,SDN08}). 
There has recently been a great deal of interest
in solving grasping problems
using partially observable MDPs (POMDPs). For example, 
Hsiao et al. \citeyear{HKL07} demonstrated  how 
simple grasping problems could be solved using POMDPs.
}%

\commentout{
There has also been some work on autonomous car drifting. 
Kolter et al.~\citeyear{KPJNT10} provide
results on 
autonomous car parking with drifting actions using a probabilistic method that
combines open-loop and closed-loop control. However, their
emphasis is more on exact trajectory-following---making a real vehicle 
accurately follow a single sliding trajectory provided by a human
expert; they do not explore the learning of general autonomous drifting
with a wide range 
of turning angles. }
\noindent {\bf Acknowledgments:} The work of Halpern and Rong was
supported in part by NSF grants 
IIS-0911036 and CCF-1214844, and by AFOSR grant
FA9550-12-1-0040, and ARO grant W911NF-14-1-0017.
The work of Saxena was supported in part by  
a
Microsoft Faculty Fellowship.

\bibliographystyle{chicago}
\bibliography{nan}  

\begin{thebibliography}{}

\bibitem[\protect\citeauthoryear{Abbeel and Ng}{Abbeel and Ng}{2005}]{AN05ICML}
Abbeel, P. and A.~Y. Ng (2005).
\newblock Exploration and apprenticeship learning in reinforcement learning.
\newblock In {\em Proc.~22nd Int. Conf. on Machine Learning (ICML '05)}, pp.\
  1--8.

\bibitem[\protect\citeauthoryear{Antos and Munos}{Antos and Munos}{2007}]{AM07}
Antos, A. and R.~Munos (2007).
\newblock Fitted {Q}-iteration in continuous action-space {MDP}s.
\newblock In {\em Advances in Neural Information Processing Systems 20 (Proc.
  of NIPS 2007)}, pp.\  9--16.

\bibitem[\protect\citeauthoryear{Brafman and Tennenholtz}{Brafman and
  Tennenholtz}{2002}]{BT02}
Brafman, R.~I. and M.~Tennenholtz (2002).
\newblock {R-MAX}: {A} general polynomial time algorithm for near-optimal
  reinforcement learning.
\newblock {\em Journal of Machine Learning Research\/}~{\em 3}, 213--231.

\bibitem[\protect\citeauthoryear{Budden, Walker, Flannery, and Mendes}{Budden
  et~al.}{2013}]{BWFM2013}
Budden, D., J.~Walker, M.~Flannery, and A.~Mendes (2013).
\newblock Probabilistic gradient ascent with applications to bipedal robot
  locomotion.
\newblock In {\em Australasian Conference on Robotics and Automation (ACRA
  2013)}, pp.\  37--45.

\bibitem[\protect\citeauthoryear{Cheng and Lin}{Cheng and Lin}{1997}]{CL1997}
Cheng, M.-Y. and C.-S. Lin (1997).
\newblock Genetic algorithm for control design of biped locomotion.
\newblock {\em Journal of Robotic Systems\/}~{\em 14\/}(5), 365--373.

\bibitem[\protect\citeauthoryear{Chernova and Veloso}{Chernova and
  Veloso}{2004}]{CV2004}
Chernova, S. and M.~Veloso (2004).
\newblock An evolutionary approach to gait learning for four-legged robots.
\newblock In {\em Proc.~International Conference on Intelligent Robots and
  Systems (IROS 2004) Vol. 3}, pp.\  2562 -- 2567.

\bibitem[\protect\citeauthoryear{Dean, Kim, and Givan}{Dean
  et~al.}{1998}]{DKG98}
Dean, T., K.~Kim, and R.~Givan (1998).
\newblock Solving stochastic planning problems with large state and action
  spaces.
\newblock In {\em Proc.~4th Int.~Conf.~on Artificial Intelligence Planning
  Systems}, pp.\  102--110.

\bibitem[\protect\citeauthoryear{Dearden, Friedman, and Andre}{Dearden
  et~al.}{1999}]{DFA99}
Dearden, R., N.~Friedman, and D.~Andre (1999).
\newblock Model based bayesian exploration.
\newblock In {\em Proc.~15th Conf. on Uncertainty in AI (UAI '99)}, pp.\
  150--159.

\bibitem[\protect\citeauthoryear{Feng, Dearden, Meuleau, and Washington}{Feng
  et~al.}{2004}]{FDMW04}
Feng, Z., R.~Dearden, N.~Meuleau, and R.~Washington (2004).
\newblock Dynamic programming for structured continuous {Markov} decision
  problems.
\newblock In {\em In Proc.~20th Conf.~on Uncertainty in Artificial
  Intelligence}, pp.\  154--161.

\bibitem[\protect\citeauthoryear{Gonçalves and Zampieri}{Gonçalves and
  Zampieri}{2006}]{GZ2006}
Gonçalves, J.~B. and D.~E. Zampieri (2006, 12).
\newblock {An integrated control for a biped walking robot}.
\newblock {\em {Journal of the Brazilian Society of Mechanical Sciences and
  Engineering}\/}~{\em 28}, 453 -- 460.

\bibitem[\protect\citeauthoryear{Guestrin, Patrascu, and Schuurmans}{Guestrin
  et~al.}{2002}]{GPS2002}
Guestrin, C., R.~Patrascu, and D.~Schuurmans (2002).
\newblock Algorithm-directed exploration for model-based reinforcement learning
  in factored mdps.
\newblock In {\em Proc. ~International Conference on Machine Learning (ICML
  2002)}, pp.\  235--242.

\bibitem[\protect\citeauthoryear{Halpern, Rong, and Saxena}{Halpern
  et~al.}{2010}]{HRS10}
Halpern, J.~Y., N.~Rong, and A.~Saxena (2010).
\newblock {MDP}s with unawareness.
\newblock In {\em Proc.~26th Conf. on Uncertainty in AI (UAI '10)}, pp.\
  228--235.

\bibitem[\protect\citeauthoryear{Hasegawa, Arakawa, and Fukuda}{Hasegawa
  et~al.}{2000}]{HAF2000}
Hasegawa, Y., T.~Arakawa, and T.~Fukuda (2000).
\newblock Trajectory generation for biped locomotion robot.
\newblock {\em Mechatronics\/}~{\em 10}, 67--89.

\bibitem[\protect\citeauthoryear{Hauskrecht, Meuleau, Kaelbling, Dean, and
  Boutilier}{Hauskrecht et~al.}{1998}]{NMKD98}
Hauskrecht, M., N.~Meuleau, L.~Kaelbling, T.~Dean, and C.~Boutilier (1998).
\newblock Hierarchical solution of {M}arkov decision processes using
  macro-actions.
\newblock In {\em Proc.~14th Conf. on Uncertainty in AI (UAI '98)}, pp.\
  220--229.

\bibitem[\protect\citeauthoryear{Hordijk and {Van der Duyn Schouten}}{Hordijk
  and {Van der Duyn Schouten}}{1984}]{HV84}
Hordijk, A. and F.~A. {Van der Duyn Schouten} (1984).
\newblock Discretization and weak convergence in {Markov} decision drift
  processes.
\newblock {\em Mathematics of Operations Research\/}~{\em 9}, 112--141.

\bibitem[\protect\citeauthoryear{Huang, Yokoi, Kajita, Kaneko, Arai, Koyachi,
  and Tanie}{Huang et~al.}{2001}]{HYKKAKT2001}
Huang, Q., K.~Yokoi, S.~Kajita, K.~Kaneko, H.~Arai, N.~Koyachi, and K.~Tanie
  (2001).
\newblock Planning walking patterns for a biped robot.
\newblock {\em IEEE Transactions on Robotics and Automation\/}~{\em 17},
  280--289.

\bibitem[\protect\citeauthoryear{Kajita, Kanehiro, Kaneko, Fujiwara, Harada,
  Yokoi, and Hirukawa}{Kajita et~al.}{2003}]{KKKFHYH2003}
Kajita, S., F.~Kanehiro, K.~Kaneko, K.~Fujiwara, K.~Harada, K.~Yokoi, and
  H.~Hirukawa (2003).
\newblock Biped walking pattern generation by using preview control of
  zero-moment point.
\newblock In {\em Proc.~International Conference on Robotics and Automation
  (ICRA 2003), Vol. 2}, pp.\  1620 -- 1626.

\bibitem[\protect\citeauthoryear{Kakade, Kearns, and Langford}{Kakade
  et~al.}{2003}]{KKL03}
Kakade, S., M.~Kearns, and J.~Langford (2003).
\newblock Exploration in metric state spaces.
\newblock In {\em Proc.~20th Int. Conf. on Machine Learning (ICML '03)}, pp.\
  306--312.

\bibitem[\protect\citeauthoryear{Kearns and Koller}{Kearns and
  Koller}{1999}]{KK99}
Kearns, M. and D.~Koller (1999).
\newblock Efficient reinforcement learning in factored {MDP}s.
\newblock In {\em Proc.~Sixteenth Int. Joint Conf. on Artificial Intelligence
  (IJCAI '99)}, pp.\  740--747.

\bibitem[\protect\citeauthoryear{Kearns and Singh}{Kearns and
  Singh}{2002}]{KS02}
Kearns, M. and S.~Singh (2002).
\newblock Near-optimal reinforcement learning in polynomial time.
\newblock {\em Machine Learning\/}~{\em 49\/}(2-3), 209--232.

\bibitem[\protect\citeauthoryear{Khalil}{Khalil}{2002}]{Khalil02}
Khalil, H.~K. (2002).
\newblock {\em Nonlinear Systems}.
\newblock Prentice-Hall.

\bibitem[\protect\citeauthoryear{Kim, Park, and Oh}{Kim et~al.}{2007}]{KPO2007}
Kim, J., I.~Park, and J.~Oh (2007).
\newblock Walking control algorithm of biped humanoid robot on uneven and
  inclined floor.
\newblock {\em Journal of Intelligent and Robotic Systems\/}~{\em 48\/}(4),
  457--484.

\bibitem[\protect\citeauthoryear{Kim and Uther}{Kim and Uther}{2003}]{KU2003}
Kim, M.~S. and W.~Uther (2003).
\newblock Automatic gait optimisation for quadruped robots.
\newblock In {\em Proc.~Australasian Conference on Robotics and Automation
  (ACRA)}.

\bibitem[\protect\citeauthoryear{Kohl and Stone}{Kohl and Stone}{2004}]{KS2004}
Kohl, N. and P.~Stone (2004).
\newblock Policy gradient reinforcement learning for fast quadrupedal
  locomotion.
\newblock In {\em Proc.~{IEEE} International Conference on Robotics and
  Automation (ICRA 2004)}, pp.\  2619--2624.

\bibitem[\protect\citeauthoryear{Mordatch, Mishra, Eppner, and Abbeel}{Mordatch
  et~al.}{2016}]{MMEA2016}
Mordatch, I., N.~Mishra, C.~Eppner, and P.~Abbeel (2016).
\newblock Combining model-based policy search with online model learning for
  control of physical humanoids.
\newblock Preprint,
  http://www.eecs.berkeley.edu/~igor.mordatch/darwin/paper.pdf.

\bibitem[\protect\citeauthoryear{Parker and Tarimo}{Parker and
  Tarimo}{2011}]{PT2011}
Parker, G. and W.~Tarimo (2011).
\newblock Using cyclic genetic algorithms to learn gaits for an actual
  quadruped robot.
\newblock In {\em Proc.~IEEE International Conference on Systems, Man, and
  Cybernetics (SMC 2011)}, pp.\  1938 -- 1943.

\bibitem[\protect\citeauthoryear{Parker, Tarimo, and Cantor}{Parker
  et~al.}{2011}]{PTC2011}
Parker, G.~B., W.~T. Tarimo, and M.~Cantor (2011).
\newblock Quadruped gait learning using cyclic genetic algorithms.
\newblock In {\em Proc.~IEEE Congress on Evolutionary Computation (CEC 2011)},
  pp.\  1529 -- 1534.

\bibitem[\protect\citeauthoryear{Picado, Gestal, Lau, Reis, and Tome}{Picado
  et~al.}{2009}]{PGLRT2009}
Picado, H., M.~Gestal, N.~Lau, L.~P. Reis, and A.~M. Tome (2009).
\newblock Automatic generation of biped walk behavior using genetic algorithms.
\newblock In {\em Proc.~10th International Work-Conference on Artificial Neural
  Networks}, pp.\  805--812.

\bibitem[\protect\citeauthoryear{Puterman}{Puterman}{1994}]{Puterman1994}
Puterman, M.~L. (1994).
\newblock {\em Markov Decision Processes}.
\newblock John Wiley and Sons, Inc.

\bibitem[\protect\citeauthoryear{Rachelson, Garcia, and Fabiani}{Rachelson
  et~al.}{2008}]{RGF08}
Rachelson, E., F.~Garcia, and P.~Fabiani (2008).
\newblock Extending the bellman equation for {MDP}s to continuous actions and
  continuous time in the discounted case.
\newblock In {\em 10th Int.~Symp.~on Artificial Intelligence and Mathematics}.

\bibitem[\protect\citeauthoryear{Schulman, Levine, Moritz, Jordan, and
  Abbeel}{Schulman et~al.}{2015}]{SLMJA2015}
Schulman, J., S.~Levine, P.~Moritz, M.~Jordan, and P.~Abbeel (2015).
\newblock Trust region policy optimization.
\newblock In {\em Proc.~31st International Conference on Machine Learning (ICML
  '15)}, pp.\  1889--1897.

\bibitem[\protect\citeauthoryear{Strom, Slavov, and Chown}{Strom
  et~al.}{2010}]{SSC2010}
Strom, J., G.~Slavov, and E.~Chown (2010).
\newblock Omnidirectional walking using {ZMP} and preview control for the nao
  humanoid robot.
\newblock In {\em RoboCup 2009: Robot Soccer World Cup XIII}, pp.\  378--389.

\bibitem[\protect\citeauthoryear{Sutton, Precup, and Singh}{Sutton
  et~al.}{1999}]{SPS99}
Sutton, R., D.~Precup, and S.~Singh (1999).
\newblock Between {MDP}s and semi-{MDP}s: A framework for temporal abstraction
  in reinforcement learning.
\newblock {\em Artificial Intelligence\/}~{\em 112}, 181--211.

\bibitem[\protect\citeauthoryear{Tedrake, Zhang, and Seung}{Tedrake
  et~al.}{2005}]{TZS2005}
Tedrake, R., T.~W. Zhang, and H.~S. Seung (2005).
\newblock Learning to walk in 20 minutes.
\newblock In {\em Proc.~Fourteenth Yale Workshop on Adaptive and Learning
  Systems}.

\bibitem[\protect\citeauthoryear{Xue, Chen, Liu, and Nardi}{Xue
  et~al.}{2012}]{XCLN2012}
Xue, F., X.~Chen, J.~Liu, and D.~Nardi (2012).
\newblock Real time biped walking gait pattern generator for a real robot.
\newblock In T.~R{\"o}fer, N.~M. Mayer, J.~Savage, and U.~Saranl{\i} (Eds.),
  {\em RoboCup 2011: Robot Soccer World Cup XV}, pp.\  210--221. Springer.

\bibitem[\protect\citeauthoryear{Zinober}{Zinober}{1989}]{Zin89}
Zinober, A.~S. (1989).
\newblock Deterministic control of uncertain systems.
\newblock In {\em Proc.~Int.~ Conf.~on Control and Applications (ICCON '89)},
  pp.\  645--650.

\bibitem[\protect\citeauthoryear{Zykov, Bongard, and Lipson}{Zykov
  et~al.}{2004}]{ZBL2004}
Zykov, V., J.~Bongard, and H.~Lipson (2004).
\newblock Evolving dynamic gaits on a physical robot.
\newblock In {\em Proc.~Genetic and Evolutionary Computation Conference, Late
  Breaking Paper (GECCO'04)}.

\end{thebibliography}
\normalsize

\fullv{
\section*{Appendix}
\commentout{
\subsection{Proof of Theorem \ref{t11}}
\prf
(a) We first consider the case where $K$ has infinitely many members. 

Consider the sum of $D(1,t)$ at each discretization level. Denote the sum for level $l$ as $SD_l$.
At discretization level 1, there are $n$ possible actions. The brute-force 
method explores the actions systematically.
$$\begin{array}{lll}
SD_1 &=& \sum^{n-1}_{t=0}\frac{q_1}{n-tq_1}\\
\end{array}$$
\noindent
Discretization level 2:\\
$$\begin{array}{lll}
SD_2 &=& \sum^{n^2-1}_{t=0}\frac{\frac{q_2}{n^2}}{1-q_1-\frac{t}{n^2}q_2}\\
														&=& \sum^{n^2-1}_{t=0}\frac{q_2}{(1-q_1)n^2-tq_2}
\end{array}$$
Note that we normalize the probability value over the sum of remaining probabilities.
This is because $D(1,t)$ is a conditional probability: it is the probability of discovering a new action \textit{given that} no action was discovered during the previous $t-1$ times. 

\noindent
Discretization level $L$:\\
$$\begin{array}{lll}
SD_L &=& \sum_{t=0}^{n^L-1} \frac{q_L}{n^L(1-\sum_1^{L-1} q_j)-tq_L}
\end{array}$$

Since 
$$\begin{array}{lll}
	\int\frac{q_L}{n^L(1-\sum_1^{L-1} q_j)-tq_L} dt= -\ln(n^L(1-\sum_1^{L-1}q_j)-tq_L),
\end{array}$$
thus,
$$\begin{array}{lll}
SD_L &\ge& \int_{t=0}^{n^L-1} \frac{q_L}{n^L(1-\sum_1^{L-1} q_j)-(t-1)q_L} dt \\
		&=& \ln \frac{n^L(1-\sum_1^{L-1}q_j)+q_L}{n^L(1-\sum_1^{L}q_j)+q_L}\\
\end{array}$$
Consider the following two cases:
\begin{itemize}
	\item case 1: if $q_L\ge 2(1-\sum^L_1 q_j)$, then we have
$$\begin{array}{lll}
\ln \frac{n^L(1-\sum_1^{L-1}q_j)+q_L}{n^L(1-\sum_1^{L}q_j)+q_L}
		&\ge& \ln \frac{n^L q_L+q_L}{\frac{n^L}{2}q_L+q_L}\\
		&=& \ln \frac{n^L+1}{\frac{n^L}{2}+1}\\
		&\ge& \ln \frac{4}{3}
\end{array}$$
The first inequality is true since $\sum^\infty_1 q_j=1$, thus $(1-\sum_1^{L-1}q_j)\ge q_L$. The third line is true for all $n, L\ge 1$.
	
	\item case 2: if $q_L< 2(1-\sum^L_1 q_j)$,
$$\begin{array}{lll}
\ln \frac{n^L(1-\sum_1^{L-1}q_j)+q_L}{n^L(1-\sum_1^{L}q_j)+q_L}
		&\ge& \ln \frac{n^L(1-\sum_1^{L-1}q_j)}{(n^L+2)(1-\sum_1^{L}q_j)}.
\end{array}$$
\end{itemize}

Now consider the accumulated sum of $D(1,t)$ for discretization level 1 to $\infty$. If \textit{case 1} holds for infinitely many $L$, then $SD_L\ge \ln\frac{4}{3}$ for infinitely many $L$, where $\ln\frac{4}{3}$ is a positive constant. Thus, $\sum_1^L SD_l=\infty$.  

Otherwise, \textit{case 1} only holds for finitely many $L$. In other words, there exists constant $t>0$ such that \textit{case 2} holds for all $L\ge t$. Thus, for $L\ge t$, we have
$$\begin{array}{lll}
	\sum_1^L SD_l &\ge& \sum_t^L SD_l\\ 
	  &\ge& \sum_{l=t}^L \ln \frac{n^l(1-\sum_1^{l-1}q_j)}{(n^l+2)(1-\sum_1^{l}q_j)}\\
	  &=& \ln (\prod_t^L \frac{1-\sum_1^{l-1}q_j}{1-\sum_1^{l}q_j})+\sum_{t}^L\ln\frac{n^l}{n^l+2}\\
		&=& \ln (\frac{1-\sum_1^{t-1}q_j}{1-\sum_1^{t}q_j}\times \frac{1-\sum_1^{t}q_j}{1-\sum_1^{t+1}q_j} \times\cdots\times 		
			\frac{1-\sum_1^{L-1}q_j}{1-\sum_1^{L}q_j})\\
		&&	\ \ +\sum_{t}^L\ln\frac{n^l}{n^l+2}\\
		&=& \ln \frac{1-\sum_1^{t-1}q_j}{1-\sum_1^{L}q_j}+\sum_{l=t}^L\ln\frac{n^l}{n^l+2}\\
\end{array}$$

Since $\sum_1^\infty q_j = 1$, when $L$ goes to infinity, $\ln \frac{1-\sum_1^{t-1}q_j}{1-\sum_1^{L}q_j}=\infty$. Furthermore, $\sum_{l=t}^L\ln\frac{n^l}{n^l+2}\ge -\frac{4}{(n-1)n^{t-1}}$, a constant. This is true since $\ln\frac{n^l}{n^l+2}\ge -2(\frac{2}{n^l+2})\ge -\frac{4}{n^l}$ for all $n, l\ge 1$. %
Thus, $\Psi(\infty)=\sum_1^\infty SD_l$ diverges. 

(b) Now consider when there are only finitely many members in $K$.
Since there are only finite possible levels, 
it is clear that at the last level, $D(1,t)$ will
become 1 eventually, and the DM is guaranteed to obtain an $\epsilon$-near-optimal policy.
\eprf
}

In this appendix, we provide proofs of Theorems~\ref{t11} and
\ref{t11}.  We start with Theorem~\ref{t12}.  We repeat the statement of
the theorem for the reader's convenience.

\othm{t12}
Using an exploration method where $D_i(1,t)\ge \beta$ for all $i,t>0$ where
$\beta\in(0,1)$ is a constant,  
for any $\alpha>0$ and $0<\delta<1$, the robot can obtain an 
$\alpha$-optimal policy to $M_{\infty}$ with probability at least $1-\delta$
in time polynomial in $l$, $|A_l|$, $|S_l|$, $1/\beta$, $1/\alpha$,
$1/\delta$, $R_{\max}$ and $T^l$,  
where $l$ is the smallest $i$ such that the optimal policy for $M_i'$ is
$(\alpha/2)$-optimal 
to $M_\infty$, 
$R_{\max}^l$ is the maximum reward (that a transition can obtain) in $M_l'$,
and $T^l$ is the $\epsilon$-return mixing time for $M_l'$.
\eothm

\begin{figure}
\centering
\includegraphics[width=.6\linewidth]{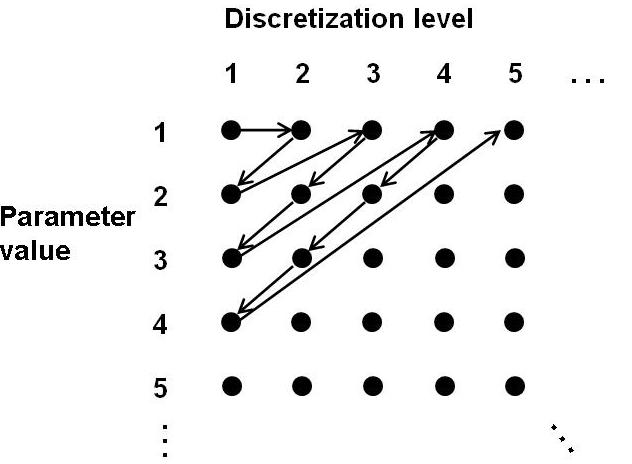}
\caption{The diagonal execution of URMAX.}
\label{fig:diagonal}
\end{figure}

\prf
Since $D_i(1,t)\ge \beta$ for $i\ge 1$, for all
levels $i\ge 1$, we have $\Psi(t)\ge \beta\cdot t\ge \beta\ln(t)$ for all
$t\ge 1$. 
By Theorem \ref{cor:poly}, for each level $i$, with probability at least
$1-\delta$, we can obtain a policy $\pi_i$ that is $(\alpha/2)$-optimal
for $M_i'$ in time polynomial in $|A_i|$, $|S_i|$, $1/\beta$, $2/\alpha$, 
$1/\delta$, $R_{\max}^i$, and $T^i$ using the URMAX algorithm,
where 
$R_{\max}^i$ is the maximum reward in $M_i'$ and 
$T^i$
is the $(\alpha/2)$-return mixing time for $M_i'$. 
However, we are not interested in obtaining a near-optimal policy for
$M_i'$; we want a near-optimal policy for $M_\infty$.
Let $l$ be the smallest
$i$ such that the optimal policy for $M_i'$ is  
$(\alpha/2)$-optimal for $M_\infty$. 
Such a level $l$ must exist, due to the continuity of the reward and
transition functions of $M_\infty$.  For suppose that $\pi$ is the
optimal policy in $M_\infty$.
By continuity, there exists a level $l$
and a policy $\pi'$ at level $l$ such that the expected reward of
$\pi'$ is within $\alpha/2$ of that of $\pi$.  But then the optimal
policy at level $l$ must have expected reward within $\alpha/2$ of that
of $\pi$.%
\footnote{Actually, there may not be an optimal policy in $M_\infty$.
That is, there may be a reward $\gamma$ such that no policy in $M_\infty$
has a reward of $\gamma$ or greater, and a sequence of 
policies $\pi^1, \pi^2, \ldots$ such that the expected reward of the
$\pi^j$ approaches $\gamma$, although no policy in the sequence actually
has expected reward $\gamma$.  But essentially the same argument still
applies: We take a policy $\pi$ in $M_\infty$ that has a reward greater than
$\gamma - \alpha/4$ and and choose a level $l$ that has a policy
approximating $\pi$ within $\alpha/4$.}
Since $\pi_l$ is $(\alpha/2)$-optimal for $M_l'$, and the optimal policy
for $M_l'$ is $(\alpha/2)$-optimal for $M_\infty$, $\pi_l$ must be
$\alpha$-optimal for $M_\infty$. 
Thus, if we knew $l$ and the values of all the relevant parameters, then
by running URMAX at each level 
from 1 to $l$, we could obtain an $\alpha$-optimal policy for $M_\infty$
in time polynomial in $l$, $|S_l|$, $1/k$, $2/\epsilon$, $1/\delta$,
$R_{\max}^j$, $|A_j|$ and $T^j$ for all $j\in [1,l]$. 
Note that we include $|A_j|$ and $T^j$ for all $j\in[1,l]$ here. This is
because we did not 
assume each discretization level $j+1$ is a refinement of the
discretization level $j$; 
thus, it could happen that 
$R_{\max}^l<R_{\max}^j$ for some $j<l$, or 
$|A_l|<|A_j|$ for some $j<l$, or that
$T^l<T^j$ for some $j<l$. 
However, we show below that we actually need only $R_{\max}^l$, $|A_l|$, and $T^l$
(instead of all $R_{\max}^j$, $|A_j|$, and $T^j$). 
The problem with the approach described in the previous paragraph is
that we do not know $l$ nor the values 
$|S_l$, $|A_j|$, $T^j$ and $R_{\max}^j$ for $j \in [1,l]$.  We solve this
problem just as HRS solved the analogous problem when running the URMAX
algorithm (see \cite{HRS10}): we diagonalize.  Specifically, 
we start running URMAX at
discretization level 1 under the assumption that 
the parameters 
($|A_1|$, $|S_1|$, $R_{\max}^1$, $|S_1|$) all have value 1.
level 2 with the parameters set to 1, and run URMAX at level 1 with
parameters set to 2;  
and so on. The process is described in Figure \ref{fig:diagonal}.
A similar approach is used by HRS to increase the value of unknown parameters.
We call running URMAX at a specific particular discretization level
using a particular setting of the parameters an \emph{iteration} of
URMAX. For example, running URMAX at discretization level 3 with
the parameters set to 4 is one iteration of URMAX.

To deal with the fact that we do not know $l$, we always keep track of
the current candidate for best policy.  (That is, the policy that is
optimal given the current set of assumptions about values of the
parameters, given the actions that have been discovered and our current
estimate of the transition probabilities.)  
At the end of each iteration of URMAX 
we run the current candidate optimal policy a sufficient number of times
so as to guarantee that the average payoff of URMAX is close to optimal,
if the current candidate optimal policy is indeed $\alpha$-optimal.%
\footnote{The number of times that we need to run the policy is computed
in \cite{HRS10}.}
Eventually, URMAX will reach a stage where it is exploring
discretization level $l$ using values for the parameters $|A_l|$, 
$|S_l|$, 
$R_{\max}^l$
and $T^l$ that are at least as
high as the actual values.  At that point, the candidate for
optimal policy is almost certain to be  $\alpha$-optimal for $M_\infty$.
(Recall that $l$ is defined to be the discretization level at which the
optimal policy is $\alpha/2$-optimal for $M_\infty$.)
After this point, we always run a policy that is at least 
$\alpha$-optimal to $M_\infty$. 
Note that this happens even if we do not
know the value of $l$ or any of the relevant parameters.  

Thus, although URMAX runs forever, from some point in time
it has discovered an $\alpha$-optimal algorithm.
Moreover, due to the ``diagonal'' manner in which URMAX is run, the
candidate optimal policy is (with probability $1-\delta$)
$\alpha$-optimal after time 
polynomial in $l$, $|A_l|$,
$|S_l|$, $T^l$, $R_{\max}^l$, $k$, $1/\delta$, and $2/\alpha$.  From that
point on, we are guaranteed to run policies that
are $\alpha$-optimal for $M_\infty$. 
\eprf %

\othm{t11}
Using brute-force exploration, 
for any $\alpha>0$ and $0<\delta<1$, a DM can find an 
$\alpha$-optimal policy in $M_{\infty}$ with probability at least
$1-\delta$ 
in time polynomial in $l$, $|A_l'|$, $|S_l|$, $1/\alpha$, $1/\delta$, $R_{\max}^l$ and $T^l$, 
where $l$ is the least $i$ such that the optimal policy for $M_i'$ is
$(\alpha/2)$-optimal 
for $M_\infty$,
$R_{\max}^l$ is the maximum reward (that a transition can obtain) in $M_l'$,
and $T^l$ is the $\epsilon$-return mixing time for $M_l'$.
\eothm

\shortv{
\prf
The proof is similar to the proof for Theorem \ref{t12}, except that we
use brute force exploration instead of apprenticeship
exploration. 
As in the proof of Theorem \ref{t12}, we apply URMAX in a
diagonal manner. We leave details to the reader. 
\eprf
}
\fullv{
\prf
At any discretization level, there are only finitely many possible
actions.  Since the brute force exploration examines all possible
actions at each level, it is guaranteed to find all useful actions, and
thus the near-optimal policy for that level. 

We apply the URMAX algorithm diagonally as described in the Proof for
Theorem \ref{t12}, so that sooner or later, we will reach the
discretization level $i$ such that the optimal policy for $M_i'$ is
$\epsilon$-close to $P^*$, and run URMAX in that level with parameters
values no less than the real values of $|A_i|$, $|S_i|$, $T_i$ and
$R_{\max}^i$. 
Thus, we are guaranteed to obtain an $\epsilon$-near-optimal
policy. 
\eprf
}
\commentout{
\subsection{Proof of Theorem \ref{t12}}
\prf
As in the proof of Theorem \ref{t11}, the accumulated sum of
$D(1,t)$ for discretization level $L$ is
$$\begin{array}{lll} 
SD_L &=& \sum_{t=0}^{n^L-1} \frac{q_L}{n^L(1-\sum_1^{L-1} q_j)-tq_L},
\end{array}$$

When the DM has to stop at level $l<L$, we have,
$$\begin{array}{lll} 
\sum^{l}_{i=1} SD_i &=& \sum_{i=1}^l (\sum_{t=0}^{n^i-1} \frac{q_i}{n^i(1-\sum_1^{i-1} q_j)-tq_i}) < \infty.
\end{array}$$
This is true since we are summing over finite terms.
Furthermore, $D(1,t)<1$ for all $t\le l$. That is, the DM is never guaranteed to 
discover new actions at any explore action, up to discretization level $l$.
We can then use the standard technique we used in the proof of Theorem \ref{t0} to
show that there exists a constant $c$, such that no algorithm can obtain
within $c$ of the optimal reward. We leave the details to the full
paper. 
\eprf
}%
}%
\end{document}